\documentclass{article}
\usepackage{ijcai24}

\usepackage{times}
\usepackage{soul}
\usepackage{url}
\usepackage[hidelinks]{hyperref}
\usepackage[utf8]{inputenc}
\usepackage[small]{caption}
\usepackage{graphicx}
\usepackage{amsmath}
\usepackage{amsthm}
\usepackage{amssymb}
\usepackage{booktabs}
\usepackage{algorithm}
\usepackage{algorithmic}
\usepackage[switch]{lineno}
\usepackage{enumitem}
\usepackage{bm}
\usepackage{multirow}
\usepackage{stackrel}
\usepackage{pifont}
\usepackage{caption}
\usepackage{graphicx}
\usepackage{float} 
\usepackage{subcaption}

\newtheorem{assumption}{Assumption}

\newtheorem{remark}{Remark}

\title{Boosting Single Positive Multi-label Classification with Generalized Robust Loss
}

\author{
Yanxi Chen\footnote{Joint First Authors}
\and
Chunxiao Li\footnotemark[2]\and
Xinyang Dai\footnote{Equal Contribution}\and
Jinhuan Li\footnotemark[3]\and
Weiyu Sun\footnotemark[3]\and
Yiming Wang\footnotemark[3]\and\\
Renyuan Zhang\footnotemark[3]\and
Tinghe Zhang\footnotemark[3]\And
Bo Wang\footnote{Corresponding Author}\\
\affiliations
University of International Business and Economics
\emails
\{202310411355,202310411351,wangbo\}@uibe.edu.cn
}

\begin{document}

\maketitle

\begin{abstract}
Multi-label learning (MLL) requires comprehensive multi-semantic annotations that is hard to fully obtain, thus often resulting in missing labels scenarios.
In this paper, we investigate Single Positive Multi-label Learning (SPML), where each image is associated with merely one positive label.
Existing SPML methods only focus on designing losses using mechanisms such as \emph{hard} pseudo-labeling and robust losses, mostly leading to unacceptable false negatives.
To address this issue, we first propose a generalized loss framework based on expected risk minimization to provide \emph{soft} pseudo labels, and point out that the former losses can be seamlessly converted into our framework.
In particular, we design a novel robust loss based on our framework, which enjoys flexible coordination between false positives and false negatives, and can additionally deal with the imbalance between positive and negative samples.
Extensive experiments show that our approach can significantly improve SPML performance and outperform the vast majority of state-of-the-art methods on all the four benchmarks.
Our code is available at https://github.com/yan4xi1/GRLoss.
\end{abstract}

\section{Introduction}


Multi-label learning (MLL) constructs predictive models to provide multi-label assignments to unseen images
\cite{zhao2022fusion}.
Conventionally, MLL assumes that each training instance is fully and accurately labeled with all relevant classes.
However, obtaining such comprehensive labels is often expensive and even impractical in many real-world scenarios \cite{lv2019weakly}.
Due to these limitations of vanilla MLL, Multi-label Learning with Missing Labels (MLML)~\cite{wu2014multi} received broad attentions, and was usually considered as a typical weakly supervised learning problem~\cite{kim2022large}.

\nocite{zhao2022fusion}
\nocite{lv2019weakly}
\nocite{liu2021emerging}
\nocite{wu2014multi}
\nocite{kim2022large}

In this paper, we focus on an extreme variant of MLML, i.e., Single Positive Multi-label Learning (SPML), where only one label is verified as positive for a given image, leaving all the other labels as the unknown ones \cite{cole2021multi}.
Compared to a fully labeled setting, SPML poses a practically more relevant yet substantially more challenging scenario.
Since only one positive label is known, existing approaches for general MLML based on label relations, such as learning positive label correlations~\cite{huynh2020interactive}, creating label matrices~\cite{feng2020regularized}, or learning to infer missing labels~\cite{durand2019learning}, are obstructed in the context of single labeled positive~\cite{cole2021multi}.


\nocite{cole2021multi}
\nocite{huynh2020interactive}
\nocite{feng2020regularized}
\nocite{durand2019learning}

Nevertheless, in the realm of SPML, the priority still lies in effectively dealing with missing labels.
Generally speaking, two strategies are commonly leveraged.
The first strategy is to treat missing labels as unknown variables needed to be predicted~\cite{rastogi2021multi}.
As discussed above, predicting such unseen labels based on only one positive label is indeed challenging. 
The second one is to assume the missing labels are \emph{All Negative} (AN), so that to transform SPML into a fully supervised MLL problem~\cite{cole2021multi}.
To our knowledge, AN assumption is still of the most popularity for MLML, including SPML (as a baseline) \cite{liu2023revisiting}, and normally trained with Binary Cross-entropy (BCE) loss.
Following the relevant study, in this paper, AN with BCE is also considered as the baseline in our experimental part.

Unfortunately, AN assumption is oversimplified and inevitably results in a large number of false negative labels~\cite{zhang2023learning}.
Ignoring false negatives will introduce label noise, which substantially impairs SPML performance~\cite{ghiassi2023trusted}.
To circumvent the pitfalls of mislabeling in AN, pseudo-labeling has been explored~\cite{hu2021simple}.
In this work, we adapt the ideas of \emph{soft} pseudo labeling to SPML.
In addition, the remaining false negatives in pseudo-labeling are regarded as noise~\cite{liu2023revisiting}, and properly handling these noisy labels can consistently improve SPML performance~\cite{xia2023holistic}.
Therefore, we specifically design a novel robust \emph{denoising} loss to reduce the negative impact of noise in pseudo labels.

\nocite{rastogi2021multi}
\nocite{hu2021simple}
\nocite{xia2023holistic}

Another challenge in SPML is the extreme intra-class imbalance between positive and negative labels within one category, which is much severer than vanilla MLL and MLML~\cite{ridnik2021asymmetric,tarekegn2021review}.
In order to tackle intra-class imbalance in SPML, a novel approach using both entropy maximization and asymmetric pseudo-labeling has been introduced~\cite{zhou2022acknowledging}, leading to promising results.
Nevertheless, the mechanism to address the issue of positive-negative label imbalance in SPML remains relatively unexplored.
Furthermore, the imbalance across classes, \emph{a.k.a} inter-class imbalance, easily causes the Long-tailed effect~\cite{zhang2023learning}, which is even aggravated in SPML.
In this paper, we effectively mitigate both intra-class and inter-class imbalance issue by instance and class sensitive reweighting.
\nocite{tarekegn2021review}
\nocite{zhou2022acknowledging}

In this paper, aiming at solving the SPML problem, we propose a new loss function framework.
Specifically, we first derive an empirical risk estimation for SPML based on single-class unbiased risk estimation (URE), and accordingly propose a \emph{novel loss function framework} for SPML problems.
To integrate Pseudo-labeling, Loss reweighting and Robust loss into our framework, we build Generalized Robust Loss (GR Loss) for SPML, allowing previous SPML methods to be naturally incorporated.
With two mild assumptions, we offer the specific forms of the components in our GR Loss.
Extensive experimental results on four benchmark datasets demonstrate that our GR Loss can achieve state-of-the-art performance.
Our contributions can be summarized in four-fold:
\begin{itemize}[leftmargin=1em]
\item Framework level: We propose a novel SPML loss function framework to estimate the posterior probability of missing labels more accurately.
In particular, our framework can unify existing SPML methods, enabling a comprehensive and holistic understanding of these former work.

\item Formulation level: We design a tailored robust loss function for both positive and negative labels, as a novel surrogate for cross-entropy loss.
The proposed robust loss is off-the-shell and can be seamlessly plug-in for our proposed framework to build the full Generalized Robust Loss.

\item Methodology level: We elaborate the false negative and imbalance issues in SPML by gradient analysis and empirical study, and first propose a soft pseudo-labeling mechanism to address these issues. 

\item Experimental level: We demonstrate the superiority of our GR Loss by conducting extensive empirical analysis with performance comparison against state-of-the-art SPML methods across four benchmarks.
\end{itemize}

\section{Related Work}
Due to its highly incomplete labels, SPML presents unique challenge for MLL~\cite{liu2023can}, especially resulting in oversimplified prediction to falsely identify all labels as positive~\cite{verelst2023spatial}.
In addition to SPML, incomplete supervision widely occurs in the context of weakly supervised learning, where \emph{pseudo-labeling} has been widely employed~\cite{liu2022acpl,chen2023softmatch,wang2022pico,zhou2022adversarial}.
The pseudo-labeling was initially applied to SPML to waive the issue of \emph{false negatives} caused by AN assumption~\cite{cole2021multi}, where the pseudo labels were incorporated with regularization \cite{zhou2022acknowledging,liu2023revisiting}.
In particular, label-aware global consistency regularization leverages contrastive learning to extract flow structure information, so that the latent soft labels can be accurately recovered~\cite{xie2022label}.

\nocite{verelst2023spatial}
\nocite{liu2022acpl}
\nocite{liu2023revisiting}
\nocite{xie2022label}

Pseudo-labeling may introduce label noise \cite{xia2023towards,higashimoto2024unbiased}.
In order to effectively cope with noisy labels, denoising robust losses~\cite{ma2020normalized}, such as Generalized Cross Entropy (GCE)~\cite{zhang2018generalized}, Symmetric Cross Entropy (SCE))~\cite{wang2019symmetric}, and Taylor Cross Entropy~\cite{feng2021can}, have been successively developed.
In particular, robust Mean Absolute Error (MAE) loss was combined with Cross Entropy (CE) loss to simultaneously consider the robustness and generalization performance~\cite{ghosh2017robust}.
It has been shown that robust losses, typically reserved for multi-class scenario, are also effective in MLL, especially in the presence of incomplete labels~\cite{zhang2021simple,xu2022one}, where both false positive and false negative labels should be properly addressed~\cite{ghiassi2023multi}.
Our method is under the same umbrella, and specifically tailors robust loss to SPML.

\nocite{ma2020normalized}
\nocite{zhang2018generalized}
\nocite{wang2019symmetric}
\nocite{feng2021can}
\nocite{zhang2021simple}
\nocite{ghiassi2023multi}

\section{Method}
\subsection{Problem Definition}\label{define}
In partial multi-label learning \cite{xie2018partial,xie2021partial}, the full data can be represented as a triplet $(\mathbf{x}, \mathbf{y}, \mathbf{s})$,
where $\mathbf{x}\!\in\!\mathcal{X}$ represents the input feature, $\mathbf{y}\!\in\!\mathcal{Y}\!=\!{\{0,1\}}^C$ denotes the ground-truth label, and $\mathbf{s}\!\in\!\mathcal{S}\!=\!{\{0,1\}}^C$ is the observed label.
In addition, $y_{c}$ denotes the $c$-th entry of $\mathbf{y}$: $y_{c}\!=\!1$ indicates the relevance to the $c$-th class, while $y_{c}\!=\!0$ means the irrelevance.
Besides, $s_{c}$ is the $c$-th entry of $\mathbf{s}$: $s_{c}\!=\!1$ indicates the relevance to the $c$-th class, while $s_{c}\!=\!0$ means either irrelevance or relevance, i.e., the unknown.

Let $\mathcal{D}\!=\!\{(\mathbf{x}^n, \mathbf{s}^n)\}_{n=1}^{N}$ be a partially labeled MLL training set with $N$ instances.
For a given sample $\mathbf{x}^n$, we only have access to the observed label $\mathbf{s}^n$, while the ground-truth label $\mathbf{y}^n$ remains unknown.
In SPML, each instance $\mathbf{x}^n$ has only one observed positive label, leaving the observed labels for all other categories being missing, represented as $\sum_{i=1}^{C} s_{i}^n\!=\!1$.

\subsection{Expected Risk Estimation}\label{erm}
The goal of SPML is to find a function $f: \mathcal{X}\!\rightarrow\!\mathcal{Y}$ from training set $\mathcal{D}$ to predict true labels for each $\mathbf{x}\!\in\!\mathcal{X}$.
In multi-label classification, the standard approach is to independently train $C$ binary classifiers $f\!=\![f_1,f_2,\cdots,f_C]$, using a shared backbone.
For a certain class, it can be viewed as a PU-learning problem \cite{kiryo2017positive,jiang2023positive}, where $y$ and $s$ degenerate to scalars.
Let $s\!=\!1$ if $\mathbf{x}$ is labeled, and $s\!=\!0$ if it is unlabeled.
Accordingly, the surrogate expected risk function \emph{w.r.t.} loss $\mathcal{L}$ is:
\begin{equation}\label{e1}
\mathcal{R}(f_i) = \mathbb{E}_{p(\mathbf{x}, y, s)}[\mathcal{L}(f_i(\mathbf{x}), y)].
\end{equation}

Then, the empirical risk is derived from Eq.\eqref{e1}, and the details are provided in Appendix \ref{A}:
\begin{equation}\label{e2}
\begin{split}
    &N\!\cdot\!\widehat{\mathcal{R}}(f_i)=\sum_{\langle \mathbf{x}, s=1\rangle} \mathcal{L}\left(f_i(\mathbf{x}),y=1\right)\\
    &+\!\sum_{\langle \mathbf{x}, s=0\rangle} k(\mathbf{x}) \mathcal{L}\left(f_i(\mathbf{x}),y\!=\!1\right)\!+\!(1\!-\!k(\mathbf{x})) \mathcal{L}\left(f_i(\mathbf{x}),y\!=\!0\right),
\end{split}
\end{equation}
where
$k(\mathbf{x})\!=\!P(y\!=\!1 \mid \mathbf{x}, s\!=\!0)$ represents the probability that an unlabeled sample is \emph{false negative} under AN assumption.
Consequently, for the classifier $f$, the empirical risk is:
\begin{equation}\label{e3}
\begin{aligned}
N\!\cdot\!\widehat{\mathcal{R}}(f)\!=\!\sum_{n=1,i=1}^{N,C}\!s_{i}^n \mathcal{L}_{i}^{n+}\!+\!(1\!-\!s_{i}^n)\!\left[k_{i}^n \mathcal{L}_{i}^{n+}\!+\!(1\!-\!k_{i}^n) \mathcal{L}_{i}^{n-}\right],
\end{aligned}
\end{equation}
where $k_i^n\!=\!P(y_i^n\!=\!1\!\mid\!\mathbf{x}^n,s_i^n\!=\!0)$, $\mathcal{L}_{i}^{n+}\!=\!\mathcal{L}(f_i(\mathbf{x}^n),y^n_i\!=\!1)$, and $\mathcal{L}_{i}^{n-}\!=\!\mathcal{L}(f_i(\mathbf{x}^n),y^n_i\!=\!0)$, $n\!=\!1,2,\!\cdots\!,N, i\!=\!1,2,\!\cdots\!,C$.

\subsection{Novel SPML Loss Framework}
According to Eq.\eqref{e3}, we propose a novel loss function framework for SPML based on the following two considerations. 
\paragraph{Pseudo-labeling.}
Given the difficulty in evaluating accurate $k_i(\mathbf{x})$, we estimate it in an online manner: Let \( p_i(\mathbf{x})\!=\!f_i(\mathbf{x}) \) be the current model output, and we use $\hat{k}(p_i(\mathbf{x}); \beta)$ to esitimate \( k_i(\mathbf{x}) \), where \( \beta \) is the parameters.

\paragraph{Loss reweighting.}
In order to deal with the class imbalance and false negatives, we propose the class-and-instance-specific weight \( v(p_i^n; \alpha) \) to reweight different samples \emph{w.r.t.} the category, where \( \alpha \) is the parameters.
Typically, for any certain class $i$, positive and negative losses are in a unified form for \emph{balanced} classification, such as cross entropy loss \( \mathcal{L}_{i}^{+}\!=\!-\log(p_i) \), \( \mathcal{L}_{i}^{-}\!=\!-\log(1\!-\!p_i) \).
However, there is a severe intra-class and inter-class imbalance in SPML, together with noisy labels in the pseudo labels and negative samples, i.e., false negatives.
Therefore, we add \( v(p_i^n; \alpha_i) \) to the loss of each instance for each class, and use \( \mathcal{L}_{i, 1}^n, \mathcal{L}_{i, 2}^n, \mathcal{L}_{i, 3}^n \) to decouple the loss in Eq.\eqref{e3} into three surrogate losses.
Accordingly, the total loss can be expressed as:
\begin{equation}
\mathcal{L} = \frac{1}{N} \sum_{n=1}^{N} \sum_{i=1}^{C} v(p_i^n; \alpha_i) \cdot \mathcal{L}_i^n,
\end{equation}
and the class-and-instance-specific loss $\mathcal{L}_i^n$ is decoupled as:
\begin{equation}
\begin{split}
\mathcal{L}_i^n\!=\!s_{i}^n \mathcal{L}_{i, 1}^n\!+\!\left(1\!-\!s_{i}^n\right) \left[\hat{k}(p_i^n; \beta_i) \mathcal{L}_{i, 2}^n\!+\!(1\!-\!\hat{k}(p_i^n; \beta_i)) \mathcal{L}_{i, 3}^n\right].
\end{split}
\end{equation}
\subsubsection{The analysis and formulation of $\hat{k}(p;\beta)$}
In this section, we present two assumptions on $\hat{k}(p;\beta)$.
The superscript $n$ and subscript $i$ are properly omitted for simplicity.
We defer to verify their validity in Appendix \ref{E3}.
Based on these assumptions, we introduce the explicit form of the pseudo-labeling function $\hat{k}(p;\beta)$.

\begin{assumption}\label{as1}
    In the initial training phase, $\hat{k}(p;\beta)$ is nearly a constant function.
\end{assumption}
\begin{remark}\label{re1}
    In the initial phase of training, the model is unable to provide accurate prediction, thus the initial output can be approximately considered as random one.
    As a result, the probability \( \hat{k}(p;\beta) \) is independent of the output and performs as a constant function.
    Furthermore, this constant is approximately equal to the number of false-negative labels divided by the number of missing labels, i.e., \( \hat{k}(p;\beta)\!=\!\frac{\#{\{y=1,s=0\}}}{\#{\{s=0\}}} \).
\end{remark}

\begin{assumption}\label{as2}
    In the final training stage, $\hat{k}(p;\beta)$ gradually becomes a monotonically increasing function.
\end{assumption}
\begin{remark}\label{re2}
In the final stage of training, the output of the well-trained model should perfectly fit the posterior probability \( P(y\!=\!1 \mid \mathbf{x}) \).
Therefore, we have:
\begin{equation}\label{kx}
\begin{aligned}
\hat{k}(p;\beta)\!\approx\!k(\mathbf{x})=& \frac{P(y\!=\!1 \mid \mathbf{x}) \cdot P(s\!=\!0 \mid y\!=\!1, \mathbf{x})}{P(s\!=\!0 \mid x)}\\
= & \frac{P(y\!=\!1 \mid \mathbf{x}) \cdot (1\!-\!P(s = 1 \mid y\!=\!1, \mathbf{x}))}{1\!-\!P(y\!=\!1 \mid \mathbf{x}) \cdot P(s\!=\!1 \mid y\!=\! 1, \mathbf{x})}\\
= & \frac{(1\!-\!a)\cdot p}{1\!-\!a\cdot p},
\end{aligned}
\end{equation}
where we use the noise-free fact of SPML, i.e., $P(s\!=\!1\mid y\!=\!0,\mathbf{x})\!=\!0$.
Moreover, under the Selected Completely At Random (SCAR) assumption~\cite{bekker2020learning}, \( a\!=\!P(s\!=\!1 \mid y\!=\!1)\!=\!P(s\!=\!1 \mid y\!=\!1, \mathbf{x}) \) is a constant independent of the sample and ranges between 0 and 1.
As a result, $\hat{k}(p;\beta)$ is a monotonically increasing function.
\end{remark}
\paragraph{The formulation.}
According to the above two assumptions, we define the form of $\hat{k}(p;\beta)$ as Logistic function:
\begin{equation}
\hat{k}(p;\beta) = \frac{1}{1 + \exp\{- (w\cdot p + b)\}},
\label{eq:k}
\end{equation}
where the parameters \( \beta\!=\![w, b] \).
We point out that there are other choices for the form of $\hat{k}(p;\beta)$, and we use Logistic is because of its simplicity and
it can easily satisfy the requirements in the assumptions.
In detail, when $w\!=\!0$, $\hat{k}(p;\beta)$ is a constant, corresponding to the early-stage training.
On the other hand, if increasing $w\!>\!0$ and decreasing $b\!<\!0$, while maintaining their ratio within a certain range, the Logistic can reflect the characteristics of late-stage training.
In particular, as $w\!\to\!+\infty$, $\frac{b}{w}\!=\!-\tau$, $\hat{k}(p;\beta)$ turns into the following threshold based hard pseudo-labeling strategy:
\begin{equation}
\hat{k}(p;\beta) =\frac{1}{1 + \exp\{- w\cdot (p - \tau)\}}\stackrel{w\to+\infty}{\longrightarrow}
\begin{cases}
    1 & \text{if } p > \tau, \\
    1/2 & \text{if } p=\tau,\\
    0 & \text{if } p < \tau. \\
\end{cases}
\end{equation}

\paragraph{The update of \(\beta\).}
Since \(\hat{k}(p;\beta)\) is only used as a calibration for the model output \(p\) to better estimate \(k(\mathbf{x})\), we stop the gradient backpropagation of $p$ in \(\hat{k}(p;\beta)\), and instead consider \(\beta\) as a trainable parameter.
However, the performance is not satisfactory if we treat $\beta$ as parameter (see the details in the last row of Table \ref{tab:Ablation}).
Consequently, we introduce \emph{linear regularization} to deterministically update \(\beta\) along with the training process.
In other words, we assume that both \(w,b\) linearly increase with the training epochs in the following manner:
\begin{align}
&w^{(t)}=w^{(0)}+\left(w^{(T)}-w^{(0)}\right) \cdot \frac{t}{T} \label{eq:yourlabel1},\\
&b^{(t)}=b^{(0)}+\left(b^{(T)}-b^{(0)}\right) \cdot \frac{t}{T} \label{eq:yourlabel2}.
\end{align}
Here, \( t\!=\!0,1,2,\cdots,T \) denotes the current training epoch, \( T \) represents the total number of training epochs, \( w^{(0)}, b^{(0)}\) are initially fixed according to Assumption \ref{as1}, and \( w^{(T)}, b^{(T)} \) are two hyperparameters.

\subsubsection{The analysis and formulation of $v(p;\alpha)$}
When the model performs well, the output confidence \(p\) can be largely trusted.
In this circumstance, when the corresponding observed label \(s=0\), if \(p\) is close to 1, it is probably that the corresponding instance is false negative or an outlier, and in either case, its weight should be reduced.

On the other hand, inspired by focal loss \cite{lin2017focal}, in the case of imbalance between positive and negative samples, the weight of simple samples should be reasonably reduced.
Hence, when \(s\!=\!0\), if \(p\) is close to 0, the corresponding instance is probably a simple one, and its weight should be small.
Conversely, if \(p\) is close to 0.5, its weight should be larger.
Furthermore, because positive label is too limited and definitely correct, we set its weight to 1.

\paragraph{The formulation.}
Based on the above discussions, we define
\(v(p; \alpha)\) as:
\begin{equation}
\begin{aligned}
v(p ; \alpha) = \begin{cases}
    1 & s = 1, \\
    \exp\{-\frac{(p - \mu)^2}{2\sigma^2}\} & s = 0.
\end{cases}
\end{aligned}
\label{eq:yourlabel5}
\end{equation}

\paragraph{The update of $\alpha$.}
Similar to $\beta$, the parameters \( \alpha\!=\![\sigma, \mu] \) are instead set as linearly updated.
The value of \( \alpha^{(t)} \) evolve over the training epochs in the following manner:
\begin{align}
\mu^{(t)} &= \mu^{(0)} + (\mu^{(T)} - \mu^{(0)}) \cdot \frac{t}{T}, \label{eq:yourlabel3} \\
\sigma^{(t)} &= \sigma^{(0)} + (\sigma^{(T)} - \sigma^{(0)}) \cdot \frac{t}{T}, \label{eq:yourlabel4}
\end{align}
where \( \mu^{(0)}, \sigma^{(0)} \) are initially fixed, and \(\mu^{(T)}, \sigma^{(T)} \) are two hyperparameters to be determined.
\subsubsection{The analysis and formulation of $\mathcal{L}_1,\mathcal{L}_2,\mathcal{L}_3$}\label{Lq}
In order to obtain more accurate supervision information, for missing labels, we employ $\hat{k}(p;\beta)$ as a soft pseudo label to indicate the probability of its presence.
However, in the training process, we still need to deal with a large amount of noise in pseudo labels.
Training with robust loss can mitigate the noise issue.
As a contrast, binary cross-entropy loss is good at fitting but does not enjoy robustness.
In this work, we propose to combine MAE with BCE to build a novel robust loss.

Particularly, inspired by the Generalized Cross Entropy (GCE) loss, we define the loss functions as follows:
\begin{align}\label{e14}
\mathcal{L}_1 = \frac{1-p^{q_1}}{q_1}\;,
\mathcal{L}_2 = \frac{1-p^{q_2}}{q_2}\;,
\mathcal{L}_3 = \frac{1-(1-p)^{q_3}}{q_3}.
\end{align}
Here, \( q_1, q_2, q_3 \) are hyperparameters to adjust the robustness of \( \mathcal{L}_1, \mathcal{L}_2, \mathcal{L}_3 \), which can be seen as a trade-off between MAE loss and BCE loss: When \( q_j\!=\!1 \), \( \mathcal{L}_j \) is the MAE loss; when \( q_j\!\rightarrow\!0 \), \( \mathcal{L}_j \) asymptotically becomes BCE.
The closer \( q_j \) is to 0, the lower robustness the loss exhibits, allowing faster learning convergence.
Conversely, the closer \( q_j \) is to 1, the stronger robustness the loss equipped, making it more effective against incorrect supervision.
Thus, controlling \( q_j \) can balance the convergence speed and robustness.
Detailed explanations can be found in Appendix \ref{D}.
Figure (\ref{fig:loss}) shows the loss curves of \( \mathcal{L}_3 \) for different values of \( q_3 \): When \( q_3\!=\!0.001 \), the curve approximates BCE loss, and when \( q_3\!=\!1 \), it represents MAE loss.

\subsection{Relation to Other SPML Losses}\label{relation}
In this section, we summarize the existing MLML and SPML losses and show how to integrate them into our unified loss function framework.

The core of our framework is to estimate the posterior probabilities of missing labels by adopting the label confidence \(p\).
As discussed above, \(v(p; \alpha)\) and \(\hat{k}(p; \beta)\) are both functions of \(p\), which fundamentally aligns with the essence of the pseudo-labeling method.
Transforming the pseudo-labeling method into the proposed framework, \(v(p; \alpha)\) and \(\hat{k}(p; \beta)\) can be represented as:
\begin{equation}
\hat{k}(p ; \beta)=
\begin{cases}
    1 & p \geq \tau_{1}, \\
    0 & p \leq \tau_{2}, \\
    \text{undefined} & \text{otherwise},
\end{cases}
\label{eq:yourlabel10}
\end{equation}
\begin{equation}
v(p ; \alpha)=
\begin{cases}
    1 & \text{otherwise}, \\
    0 & s=0 \text{ and } \tau_{2}<p<\tau_{1}.
\end{cases}
\label{eq:yourlabel11}
\end{equation}
Here, \(\tau_{1},\tau_{2}\) are adaptively changed with the training epoch \( t \), acting as adaptive thresholds.
Besides, in existing methods, \(\mathcal{L}_1,\mathcal{L}_2,\mathcal{L}_3\) typically employ cross-entropy loss:
\begin{equation}
\begin{aligned}
\mathcal{L}_1=-\log(p),\;\mathcal{L}_2=-\log(p),\;\mathcal{L}_3=-\log(1-p).
\end{aligned}
\label{eq:yourlabel9}
\end{equation}
We present the specific forms of the components of our framework corresponding to existing MLML/SPML methods in Table \ref{t1}.
More explanations can be found in Appendix \ref{B}.
\renewcommand{\arraystretch}{1.5}
\begin{table*}
    \centering
    \resizebox{\linewidth}{24mm}{
    \begin{tabular}{|c|c|c|c|c|c|}
        \hline
        \multirow{2}{*}{Methods} 
        & EN Loss & EM Loss & Hill Loss & Focal Margin+ SPLC & GR Loss\\
        & ~\cite{verelst2023spatial}  & ~\cite{zhou2022acknowledging} & ~\cite{zhang2021simple} & ~\cite{zhang2021simple} & (Ours)\\
        \hline
        $\hat{k}(p)$ & 
        $\begin{cases}
            0 & p \leq \tau_{1} \\
            \text{undefined} & \text{otherwise}
        \end{cases}$ & 
        $\begin{cases}
            0 & p\!\leq\!\tau_{1} \\
            1 & \text{otherwise}
        \end{cases}$ & 
        $0$ & 
        $\begin{cases}
            0 & p\!\leq\!\tau_{1} \\
            1 & \text{otherwise}
        \end{cases}$ & $\frac{1}{1 + \exp\{- (w\cdot p + b)\}}$\\
      
        \hline
        $v(p)$ & 
        $\begin{cases}
            1 & \text{otherwise} \\
            0 & s\!=\!0 \text{ and } p\!>\!\tau_{1}
        \end{cases}$ & 
        $\begin{cases}
            1 & \text{otherwise} \\
            \alpha & s\!=\!0 \text{ and } p\!>\!\tau_{1}\\
            \beta & s\!=\!0 \text{ and } p\!\leq\!\tau_{1}
        \end{cases}$ & 
        $1$ & $1$ & $\begin{cases}
    1 & s = 1, \\
    \exp\{-\frac{(p - \mu)^2}{2\sigma^2}\} & s = 0.
\end{cases}$\\
        \hline
        $\mathcal{L}_{1}$ & $-\log(p)$ & $-\log(p)$ & $-\log(p)$ & $(1-p_{m})^{\gamma} \log(p_{m})$ & $\frac{1-p^{q_1}}{q_1}$\\
        \hline
        $\mathcal{L}_{2}$ & $-\log(p)$ & $p \log(p)\!+\!(1-p) \log(1-p)$ & \text{undefined} & $(1-p_{m})^{\gamma} \log(p_{m})$ & $\frac{1-p^{q_2}}{q_2}$\\
        \hline
        $\mathcal{L}_{3}$ & $-\log(1-p)$ & $-\hat{p} \log(p)\!-\!(1-\hat{p}) \log(1-p)$ & $-(\lambda-p) p^{2}$ & $-(\lambda-p) p^{2}$ & $\frac{1-(1-p)^{q_3}}{q_3}$\\
        \hline
    \end{tabular}
}
    \caption{Representing the existing MLML/SPML losses in our unified loss function framework.}
    \label{t1}
\end{table*}

\subsection{Gradient Analysis}\label{GA}

To better understand the performance of the proposed GR Loss, we conduct a gradient-based analysis, which is commonly used for in-depth study of loss functions \cite{ridnik2021asymmetric}.
Observing gradients is beneficial as, in practice, network weights are updated according to the gradient of loss.
For convenience, let $z\!=\!z_c$ denote the output logit of the $c$-th class for $\mathbf{x}$, and \(\mathcal{L}_{\varnothing}\!=\!\hat{k} \cdot \frac{1-p^{q_2}}{q_2}\!+\!(1\!-\!\hat{k}) \cdot \frac{1\!-\!(1\!-\!p)^{q_3}}{q_3}\) represent the loss for the unannotated label.
The gradients of $\mathcal{L}_{\varnothing}$ \emph{w.r.t.} \(z\) are given by:
\begin{equation}\label{e18}
g(p)\!=\!\frac{\partial \mathcal{L}_{\varnothing}}{\partial z}\!=\!\frac{\partial \mathcal{L}_{\varnothing}}{\partial p} \frac{\partial p}{\partial z}\!=\!(1-\hat{k})(1-p)^{q_3} \cdot p - \hat{k} \cdot p^{q_2} \cdot (1-p),
\end{equation}
where \(p\!=\!\sigma(z)\!=\!(1+e^{-z})^{-1}\) and \(\hat{k}\) is defined in Eq.\eqref{eq:k}. 

\begin{figure}[t!]
    \centering
    \begin{subfigure}[t]{0.235\textwidth}
        \centering
        \includegraphics[width=\textwidth]{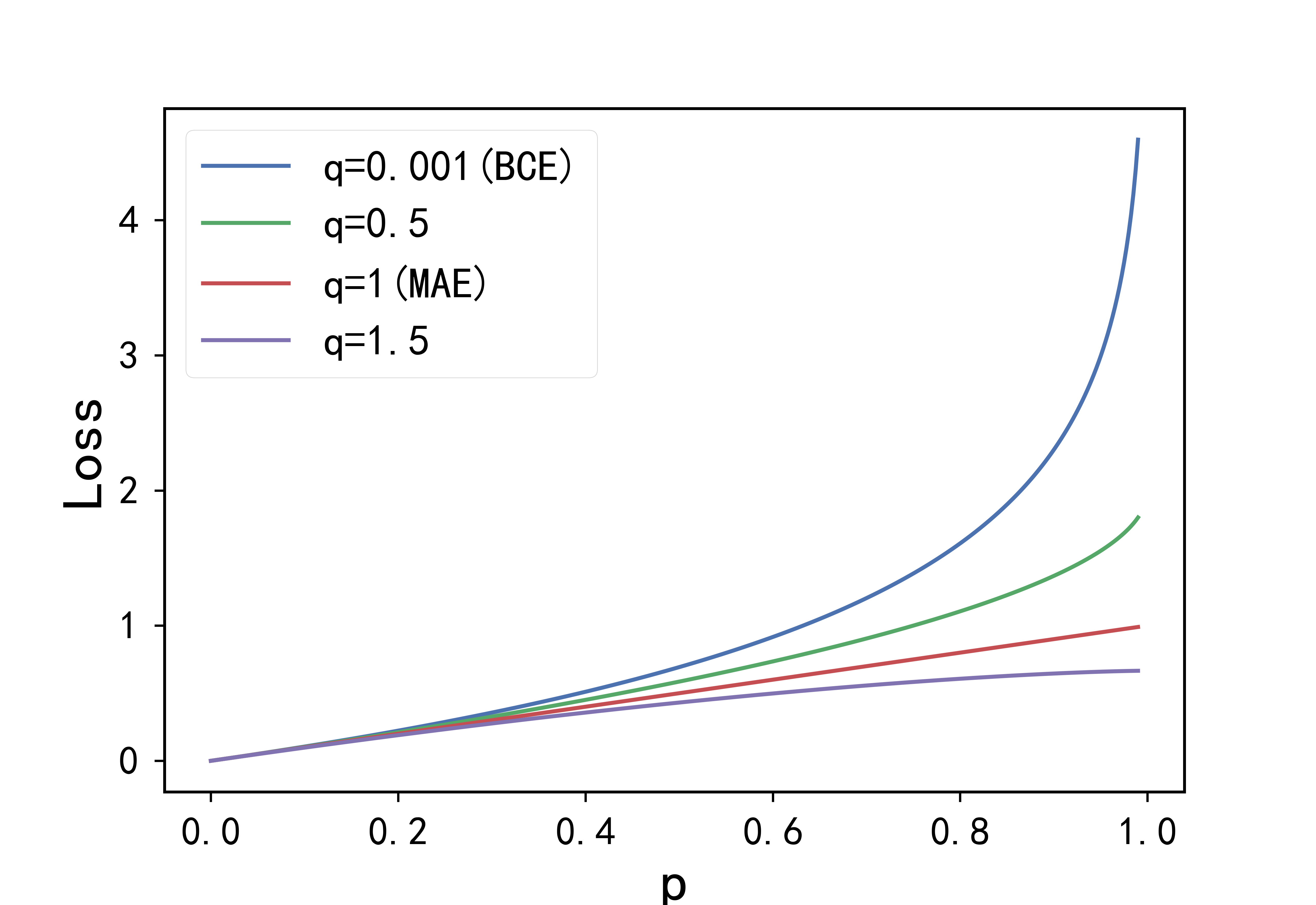}
            \caption{Loss curves for different \( q_3 \).}
            \label{fig:loss}
    \end{subfigure}
    \begin{subfigure}[t]{0.235\textwidth}
           \centering
           \includegraphics[width=\textwidth]{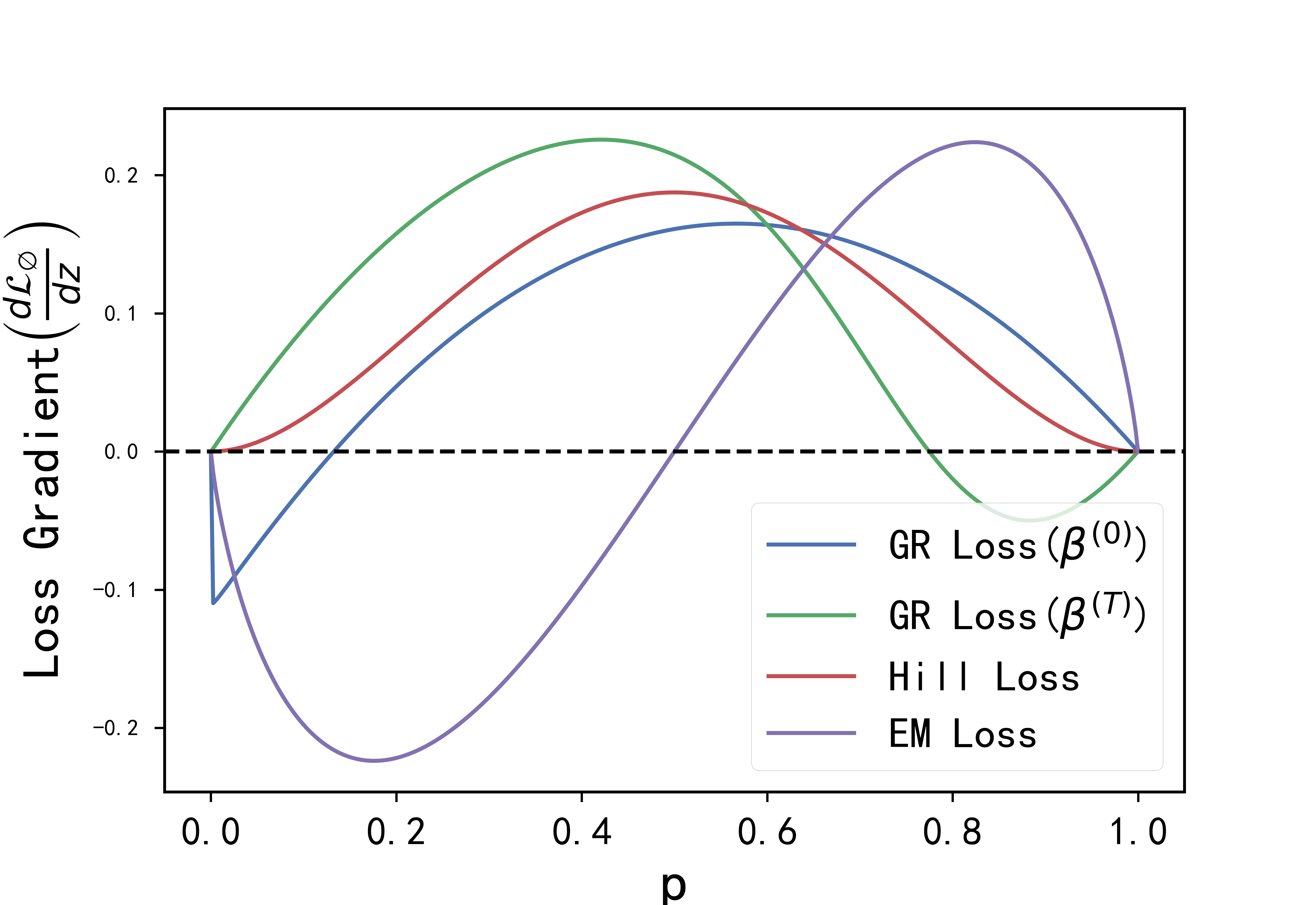}
            \caption{The gradients \( \frac{\partial \mathcal{L}_{\varnothing}}{\partial z} \) on missing label   for various losses.}
            \label{fig:gradient}
    \end{subfigure}
    \caption{The gradient analysis. Better viewed in color.}
\end{figure}

An intuitive interpretation of the gradient is: When \( g(p)>0 \), a decrease of \( p \) leads to reduction in loss; conversely, when \( g(p)<0 \), an increase of \( p \) results in reduction in loss.
Accordingly, we analyzed the gradient of GR Loss at different epochs and the loss gradients of other SPML methods, including EM loss and Hill loss.
Figure (\ref{fig:gradient}) illustrates the gradients of different losses, where \( \beta^{(0)} \) and \( \beta^{(T)} \) represent the gradients at the beginning and the end of training, respectively.
Eventually, we reach the following conclusions:
\begin{itemize}[leftmargin=1em]
\item The imbalance between positive and negative labels is a key factor to affect SPML performance.
By properly adjusting \(q_3\), we can rebalance the supervision information. 
\item The issue of false negatives is also a key factor to impact the gradient (see Eq.\eqref{e18}), which can be effectively addressed by adjusting \(\hat{k}(p;\beta)\).
\end{itemize}
A detailed derivation process and specific explanations with evidence are provided in Appendix \ref{C}.

\section{Experiments}
We provide the main empirical results of the proposed GR Loss in this section, including the comparison with state-of-the-art methods, the ablation study of our framework and robust loss, and the hyperparameter analysis.
More experimental results can be found in Appendix \ref{E}.

\renewcommand{\arraystretch}{1.25}
\begin{table*}
    \centering
    \begin{tabular}{c|c|cccc}
        \toprule
          & Methods & VOC & COCO & NUS & CUB \\
        \midrule
        \multirow{3}{*}{Baselines}     
        & AN~\cite{cole2021multi}        &85.89±0.38   &64.92±0.19   & 42.49±0.34 & 18.66±0.09 \\
        & AN-LS ~\cite{cole2021multi}   & 87.55±0.14  &  67.07±0.20 & 43.62±0.34  & 16.45±0.27 \\
        & Focal loss~\cite{lin2017focal} &87.59±0.58   &68.79±0.14   & 47.00±0.14  & 19.80±0.30 \\
        \midrule
        \multirow{5}{*}{MLML Methods}
        & Hill loss~\cite{zhang2021simple}  &87.71±0.26   &71.43±0.18   &  - &  - \\
        & SPLC~\cite{zhang2021simple}    &88.43±0.34   &71.56±0.12   & 47.24±0.17  & 18.61±0.14 \\
        & LL-R~\cite{kim2022large} &89.15±0.16   &71.77±0.15  &  48.05±0.07 & 19.00±0.02 \\
        & LL-Ct~\cite{kim2022large}  &88.97±0.15   &71.20±0.29  & 48.00±0.08  & 19.31±0.16 \\
        & LL-Cp~\cite{kim2022large}      &88.62±0.23   &70.84±0.33   & 47.93±0.01  & 19.01±0.10 \\
        \midrule
        \multirow{6}{*}{SPML Methods}
        & ROLE~\cite{cole2021multi}       &88.26±0.21   &69.12±0.13   & 41.95±0.21  & 14.80±0.61 \\
        & EM~\cite{zhou2022acknowledging}     &88.67±0.08   &70.64±0.09 & 47.25±0.30 & 20.69±0.53 \\
        & EM+APL~\cite{zhou2022acknowledging}     &89.19±0.31   &70.87±0.23   & 47.78±0.18  & 21.20±0.79 \\
        & SMILE~\cite{xu2022one}  &87.31±0.15 &70.43±0.21   & 47.24±0.17 & 18.61±0.14 \\
        & MIME~\cite{liu2023revisiting} & \underline{89.20±0.16} &  \underline{72.92±0.26} & \underline{48.74±0.43} & \textbf{21.89±0.35} \\
         &GR Loss (ours) &\textbf{89.83±0.12} 
        &\textbf{73.17±0.36}  &  \textbf{49.08±0.04} 
        & \underline{21.64±0.32} \\
        \bottomrule
    \end{tabular}
    \caption{The mean Average Precision (mAP, \%) of our GR Loss and the comparing methods on four SPML benchmarks.
    The best performance of the methods are marked in bold and the second best are marked with underlines.}
    \label{tab:plain}
\end{table*}

\subsection{Experimental Setup}
\paragraph{Dataset.}
We evaluate our proposed GR Loss on four benchmark datasets: Pascal VOC-2012 (VOC)~\cite{everingham2012pascal}, MS-COCO-2014(COCO)~\cite{lin2014microsoft}, NUS-WIDE(NUS)~\cite{chua2009nus}, and CUB-200-2011(CUB)~\cite{wah2011caltech}.
We first simulate the single-positive label training environments commonly used in SPML~\cite{cole2021multi}, and replicate their training, validation and testing samples.
In these datasets, only one positive label is randomly selected for each training instance, while the validation and test sets remain fully labeled.
More dataset descriptions are provided in Appendix \ref{E1}.
\paragraph{Implementation details and hyperparameters.}
For fair comparison, we follow the mainstream SPML implementation ~\cite{cole2021multi}.
In detail, we employ ResNet-50~\cite{he2016deep} architecture, which was pre-trained on ImageNet dataset~\cite{russakovsky2015imagenet}.
Each image is resized into $448\!\times\!448$, and performed data augmentation by randomly flipping an image horizontally.
We initially conduct a search to determine and fix the hyperparameters $q_2$ and $q_3$ in Eq.\eqref{e14}, typically 0.01 and 1, respectively.
Because the robust loss has a significant impact on training, which is also reflected in Table \ref{tab:Ablation}.
Therefore, we only need to adjust four hyperparameters in $(\beta^{(T)}$, $\alpha^{(T)})$.
More details about hyperparameter settings are described in Appendix \ref{E2}.
\paragraph{Comparing methods.}
In our empirical study, we compare our method to the following state-of-the-art methods: AN loss (assuming-negative loss)~\cite{cole2021multi}, AN-LS (AN loss combined with Label Smoothing)~\cite{cole2021multi}, Focal loss~\cite{lin2017focal}, ROLE (Regularised Online Label Estimation)~\cite{cole2021multi}, Hill loss~\cite{zhang2021simple}, SPLC (Self-Paced Loss Correction)~\cite{zhang2021simple}, EM (Entropy-Maximization Loss)~\cite{zhou2022acknowledging}, EM+APL (Asymmetric Pseudo-Labeling)~\cite{zhou2022acknowledging}, SMILE (Single-positive
MultI-label learning with Label Enhancement)~\cite{xu2022one}, Large
Loss (LL-R, LL-Ct, LL-Cp)~\cite{kim2022large}, and MIME~\cite{liu2023revisiting}. 
The goal of all the above methods is to propose new SPML loss function, which is consistent with ours.
Indeed, we are also concerned that there are methods that adopt pre-training strategies or use different backbones, such as: LAGC~\cite{xie2022label}, DualCoop~\cite{sun2022dualcoop}, HSPNet~\cite{wang2023hierarchical}, CRISP~\cite{liu2023can}.
Due to the length limitation and in order to conduct a fair comparison, we only consider the former loss function modification approach in this study.

\nocite{everingham2012pascal}
\nocite{lin2014microsoft}
\nocite{chua2009nus}
\nocite{wah2011caltech}
\subsection{Results and Discussion}
The experimental results of most existing MLML and SPML methods on four SPML benchmarks are reported in Table \ref{tab:plain}.
It is observed that AN loss performs the worst almost on all four datasets, indicating that the false negative noise introduced by AN assumption has a significant negative impact on SPML.
Meanwhile, Focal Loss, as a baseline method, is ineffective for imbalance in SPML, especially the inter-class imbalance.

Notably, our GR Loss outperforms existing methods in the first three SPML benchmarks, i.e., VOC, COCO, NUS, and achieves the second-highest result on CUB, slightly lower than MIME.
The main reason for the second best on CUB is, our method treats SPML as \( C \) multiple binary classification tasks and the more categories means the more positive labels for each image, thus the stronger correlation between labels, leading to inferior performance of our method.
In detail, for VOC (20 classes), the mAP is 0.63\% higher than the second-best, and on COCO (80 classes) and NUS (81 classes), the improvements are 0.25\% and 0.34\%, respectively.
However, there are 312 classes in CUB and an average of 31.5 positive classes per image.
As a result, our method is slightly worse than MIME, which considers label correlation and inter-class differences.

\begin{figure*}[t!]
    \begin{subfigure}[t]{0.34\textwidth}
           \centering
           \includegraphics[width=\textwidth]{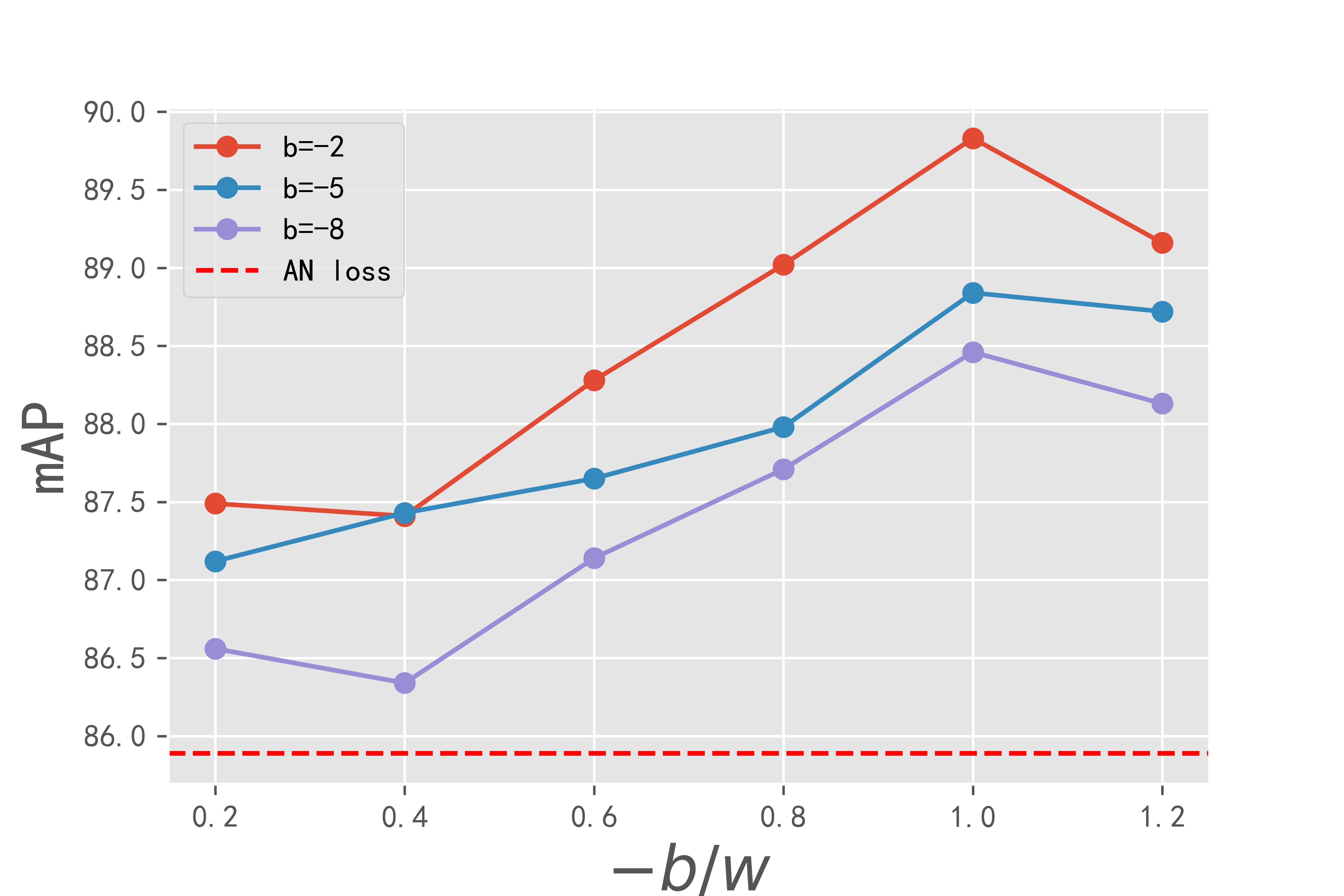}
            \caption{Analysis on $\beta$.}
            \label{fig:beta}
    \end{subfigure}
    \begin{subfigure}[t]{0.34\textwidth}
            \centering
            \includegraphics[width=\textwidth]{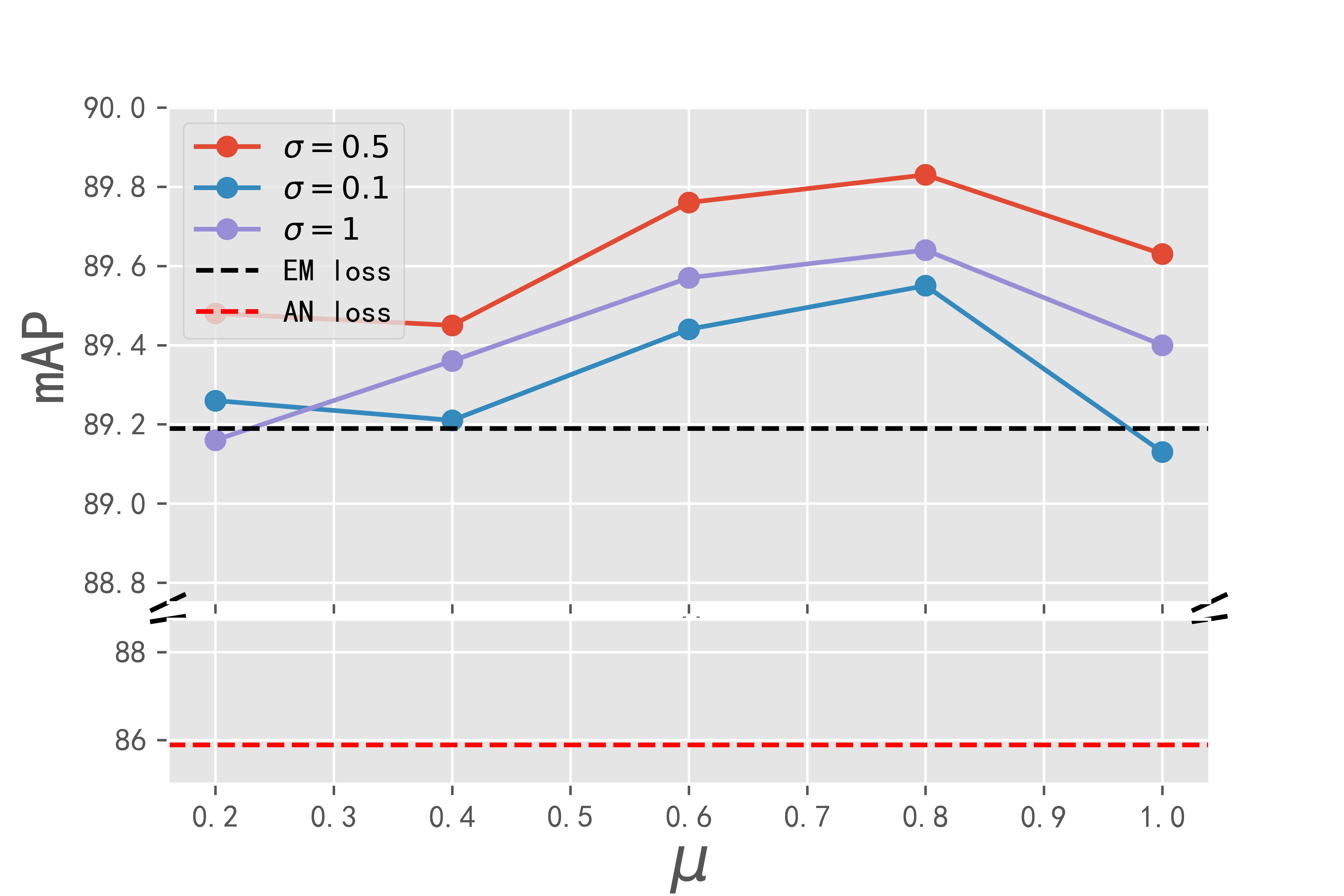}
            \caption{Analysis on $\alpha$.}
            \label{fig:alpha}
    \end{subfigure}
    \begin{subfigure}[t]{0.34\textwidth}
            \centering
            \includegraphics[width=\textwidth]{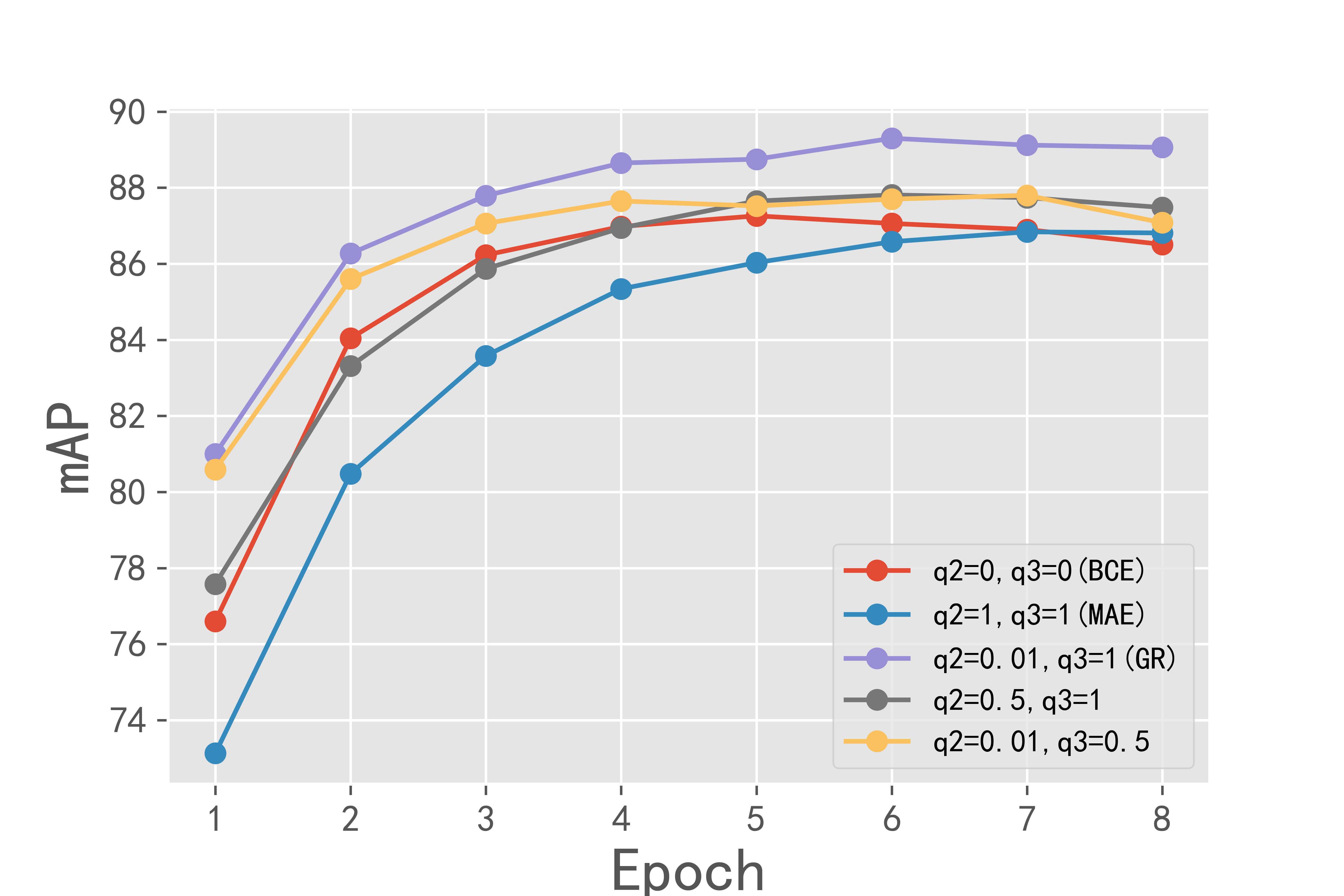}
            \caption{Analysis on $q_2$ and $q_3$.}
            \label{fig:q}
    \end{subfigure}
    \caption{The mAP (\%) of GR Loss with different hyperparameters on VOC.
    Better viewed on screen and in color.}
\end{figure*}

\begin{figure*}[t!]
    \begin{subfigure}[t]{0.35\textwidth}
           \centering
           \includegraphics[width=\textwidth]{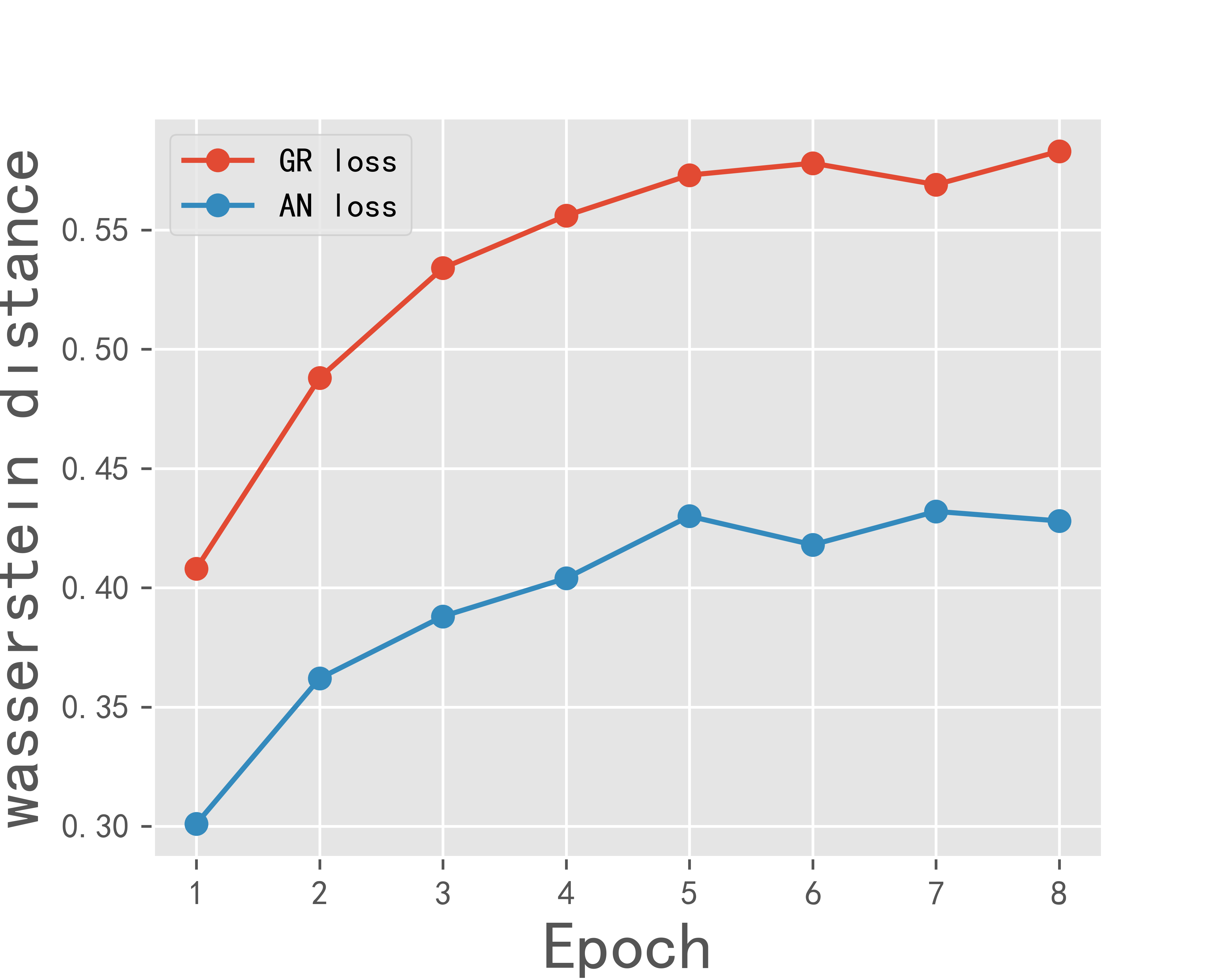}
            \caption{The Wasserstein distance for GR Loss and AN loss at each epoch on VOC.}
            \label{fig:distance}
    \end{subfigure}
    \begin{subfigure}[t]{0.33\textwidth}
           \centering
           \includegraphics[width=\textwidth]{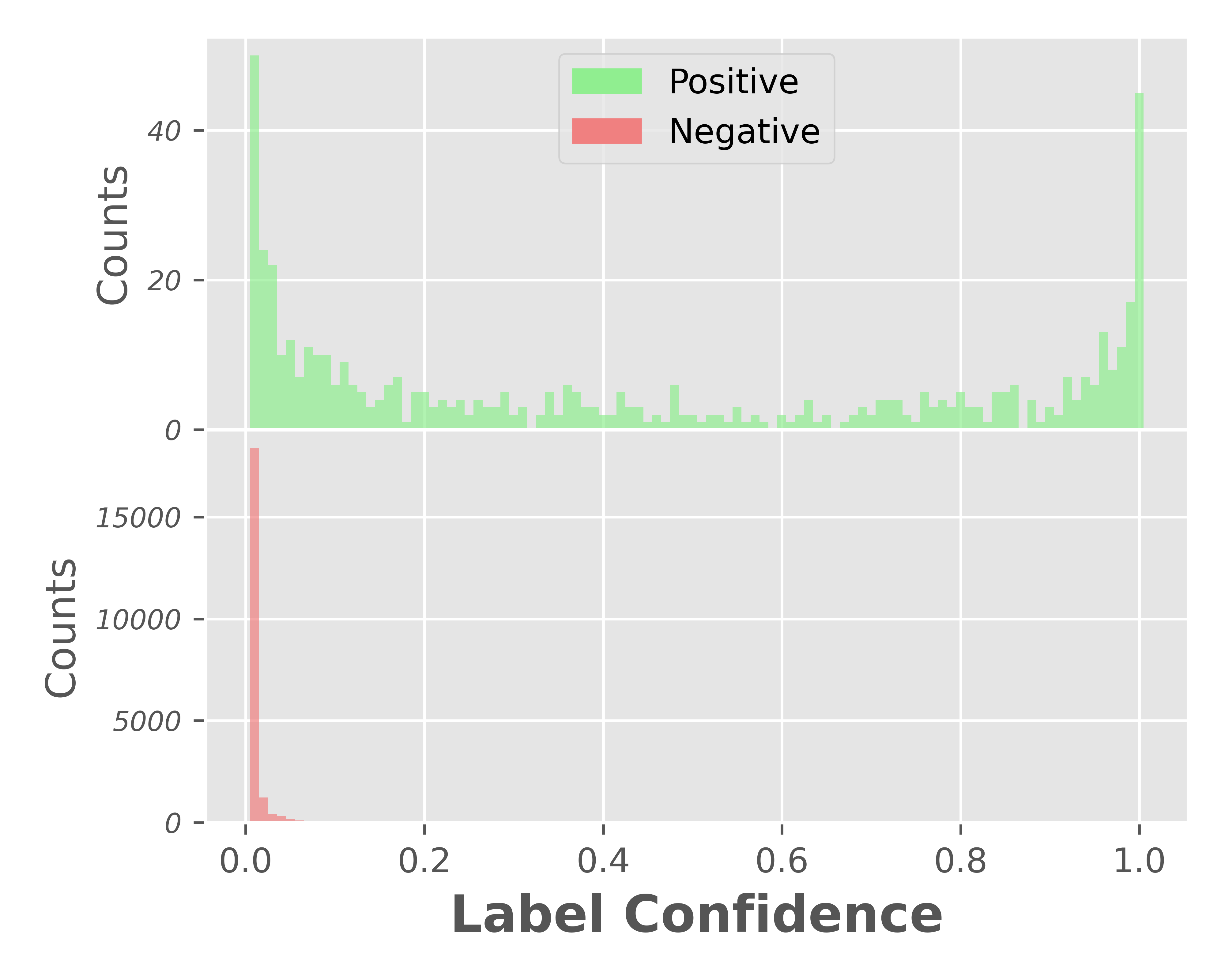}
            \caption{The confidence distribution for the predicted missing labels at the epochs where AN loss performs best.}
            \label{fig:AN}
    \end{subfigure}
    \begin{subfigure}[t]{0.33\textwidth}
            \centering
            \includegraphics[width=\textwidth]{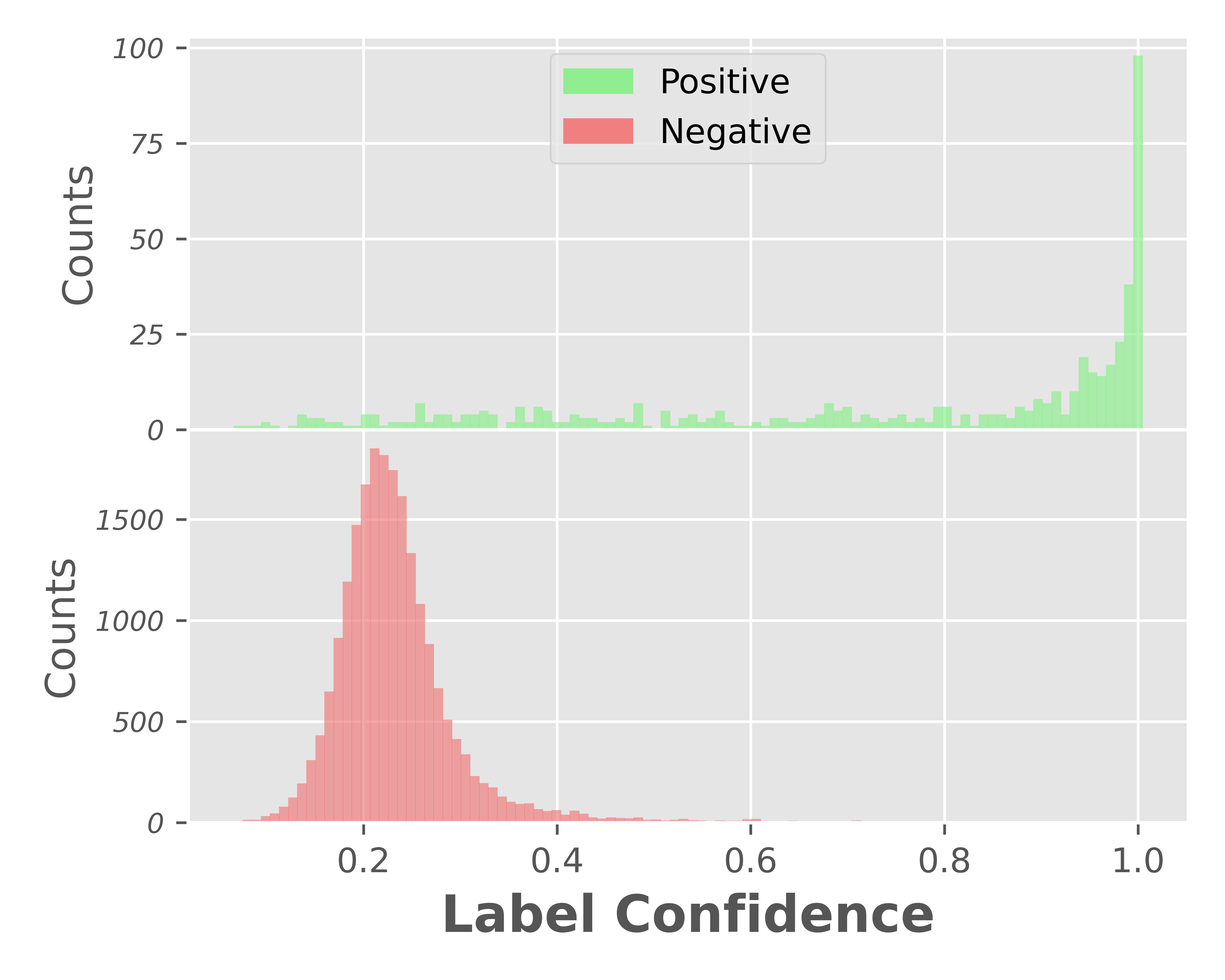}
            \caption{The confidence distribution for the predicted missing labels at the epochs where our GR Loss performs best.}
            \label{fig:gr}
    \end{subfigure}
    \caption{Distinguishing ability of model predictions on VOC.
    Better viewed on screen and in color.}
\end{figure*}

\subsection{Ablation Study}
In order to present an in-depth analysis on how the proposed method improves SPML performance, we conduct thorough ablation study on VOC and COCO and report the results in Table \ref{tab:Ablation}.
It is observed that using only $\hat{k}(p;\beta)$ results in the greatest improvement, almost a 1\% increase compared to using the other two items alone.
Utilizing both $\hat{k}(p;\beta)$ and $\mathcal{L}_1, \mathcal{L}_2, \mathcal{L}_3$ leads to a significant enhancement, with mAP reaching 89.35 and 72.86.
Moreover, adding $v(p;\alpha)$ can further elevate the model's performance, with mAP improving to 89.83 and 73.17, respectively.
The last row indicates that treating $\beta$ and $\alpha$ as trainable parameters obtains inferior results, which verifies our analysis.

\renewcommand{\arraystretch}{1.25} 
\begin{table}[t]
\centering
\begin{tabular}{ccc|cc}
\toprule 
& Methods & & VOC & COCO \\ 
\midrule 
\multicolumn{3}{c|}{AN loss (baseline)} & 85.89 & 64.92\\
\midrule
$\hat{k}(p;\beta)$ & $v(p;\alpha)$ & \multicolumn{1}{c}{$\mathcal{L}_1,\mathcal{L}_2,\mathcal{L}_3$} &  \multicolumn{2}{|c}{mAP (\%)}  \\ 
\midrule 
\checkmark & \ding{55} & \ding{55} & 88.31 & 70.41 \\
\ding{55} &\checkmark & \ding{55} & 87.32 & 69.17 \\
\ding{55} & \ding{55} &\checkmark & 87.48 & 69.96 \\
\checkmark &\checkmark & \ding{55} & 88.11 & 70.48 \\
\checkmark & \ding{55} & \checkmark& \underline{89.35}& \underline{72.86} \\
\ding{55} &\checkmark &\checkmark & 87.83 & 70.20 \\
\checkmark  &\checkmark &\checkmark & \textbf{89.83} & \textbf{73.17} \\
 $\circ$ & $\circ$ & \checkmark & 89.03 & 71.84 \\
\bottomrule
\end{tabular}
\caption{Ablation study on VOC and COCO.
A check mark $\checkmark$ indicates that the result with best hyperparameters.
A cross \ding{55} means $k(p)\!=\!0, v(p)\!=\!1, q_1\!=\!q_2\!=\!q_3\!=\!0.01$, accordingly, i.e., AN assumption.
We use a circle $\circ$ to represent the case that treating $\beta$ and $\alpha$ as trainable parameters and trained with gradient descent method.
The best performance on the datasets are marked in bold and the second best are marked with underlines.}
\label{tab:Ablation}
\end{table}
\subsection{Hyperparameter and Distinguishability}
\paragraph{The impact of $\beta$.}
We present the results of GR Loss with different $\beta^{(T)}$ in Figure (\ref{fig:beta}).
To facilitate the analysis, we adjust the threshold $\tau\!=\!-b/w$ while keeping $b^{(T)}$ fixed at -2, -5, and -8, respectively.
The results indicates that our model consistently achieves the optimal when $\tau\!=\!1$, achieving a best mAP at 89.83 with $b\!=\!-2$.
As the threshold decreasing, the performance gradually declines.
However, it is still superior to the baseline, i.e., vanilla AN loss (the red dotted line).
          

\paragraph{The impact of $\alpha$.}
We present GR Loss performance with different $\alpha^{(T)}$ in Figure (\ref{fig:alpha}).
By fixing $\sigma^{(T)}$ at 0.1, 0.5, and 1, and varying the mean $\mu$ from 0.2 to 1, all the three curves peak at 0.8.
Moreover, the mAPs for all parameter combinations are above 89 and mostly surpass both EM loss and AN loss, which indicating our method enjoys promising robustness.
\paragraph{The impact of $q$.}
Due to the rare correctness of the label with $s=1$, we set $q_1\!=\!0.01$ to enhance convergence speed and instead adjust $q_2$ and $q_3$.
The result in Figure (\ref{fig:q}) indicates that our model achieves the best mAP per epoch on the validation set when $q_2\!=\!0.01,q_3\!=\!1$.
However, if we increase the value of $q_2$ or decrease the value of $q_3$, the model's performance deteriorates.
This suggests that for missing labels, GR Loss has a higher tolerance for false positive noise.
Setting $q_2\!=\!0.01$ will reduce robustness but can accelerate the learning by offering large gradient.
In contrast, the model has less tolerance for false negative noise, hence setting $q_3\!=\!1$ where $\mathcal{L}_3$ becomes the MAE loss, allows greater accommodation of false negative noise.
Additionally, this setting reduces the learning effectiveness on negative supervision due to small gradient, thereby addressing the issue of imbalance between positive and negative samples.
\paragraph{Distinguishability of model predictions.}
A model with promising generalization should be capable of producing informative predictions for unannotated labels, i.e., the predicted probabilities for positive and negative labels should be clearly distinguishable for unannotated labels.
Hence, we assess the confidence of model's predictions on validation set for all classes and all unannotated labels at each epoch, and quantitatively measure the confidence distribution difference for positive and negative labels with Wasserstein distance.
As shown in Figure (\ref{fig:distance}), the Wasserstein distances for GR Loss are much greater than those for AN loss at each epoch.
Besides, as shown in Figure (\ref{fig:gr}), compared with AN loss in Figure (\ref{fig:AN}), for GR Loss, the confidence of negative labels is concentrated around 0.2, while that of positive labels is evenly distributed and becomes extremely low when smaller than 0.9, mainly clustering around 1.

\section{Conclusion}
In this paper, we proposed a novel loss function called Generalized Robust Loss (GR Loss) for SPML.
We employ a soft pseudo-labeling mechanism to compensate the lack of labels. 
Meanwhile, we specifically design a robust loss to cope with the noise introduced in pseudo labels. 
In addition, we introduce two tunable functions $\hat{k}(\cdot;\beta)$ and $v(\cdot;\alpha)$ to calibrate model output $p$, so that to simultaneously deal with the intra-class and inter-class imbalance.
Furthermore, we demonstrate the validity of GR Loss from both theoretical and experimental perspectives.
From the theoretical perspective, we derive empirical risk estimates for SPML and perform a gradient analysis.
Besides, in experiments, our method mostly achieves state-of-the-art results on all four benchmarks.

Nevertheless, there are three limitations in our study.
First, $q_1,q_2,q_3$ in the robust loss Eq.\eqref{e14} are all set as static hyperparameters.
However, it is better to set the robust loss to be dynamically varying so that to balance the fitting ability and robustness.
Second, as mentioned earlier, we did not evaluate our GR Loss on different backbones to fully validate its effectiveness.
Third, our method does not consider the correlation among classes, resulting in an inferior performance compared with MIME on CUB.
To overcome these limitations will shed light on the promising improvement of our current work.

\newpage
\bibliographystyle{named}
\bibliography{ijcai24}

\newpage
\appendix
\section{Expected Risk Estimation}\label{A}
In this section, we present the detailed derivation process of the expected risk function for $f_i$, denoted as $\mathcal{R}(f_i)$, corresponding to Section \ref{erm}.
\begin{equation}
\begin{aligned}
\mathcal{R}(f_i) = & \, E_{p(\mathbf{x}, y, s)}[\mathcal{L}(f_i(\mathbf{x}), y)] d\mathbf{x}dyds\\
= & \, \int_{\mathbf{x}, y, s} \mathcal{L}(f_i(\mathbf{x}), y) p(\mathbf{x}, y, s) d\mathbf{x}dyds\\
= & \, \int_{\mathbf{x}} p(\mathbf{x}) d\mathbf{x}\sum_{s=0}^{1} p(s \mid \mathbf{x}) \sum_{y=0}^{1} p(y \mid \mathbf{x}, s) \mathcal{L}(f_i(\mathbf{x}), y) \\
= & \, \int_{\mathbf{x}} p(\mathbf{x})d\mathbf{x}\left\{ p(s\!=\!1 \mid \mathbf{x}) \mathcal{L}(f_i(\mathbf{x}), y=1) \right. \\
& \, + p(s\!=\!0 \mid \mathbf{x}) \left[ p(y\!=\!1 \mid \mathbf{x}, s\!=\!0) \mathcal{L}(f_i(\mathbf{x}),y\!=\!1) \right. \\
& \, \left. \left. + p(y\!=\!0 \mid \mathbf{x}, s\!=\!0) \mathcal{L}(f_i(\mathbf{x}), y\!=\!0) \right]\right\}\vphantom{\int} \\
= & \, \int_{\mathbf{x}} P(\mathbf{x}, s\!=\!1) \mathcal{L}(f_i(\mathbf{x}), 1) d\mathbf{x}\\
& \, + \int_{\mathbf{x}}\left\{P(\mathbf{x}, s\!=\!0) p(y\!=\!1 \mid \mathbf{x}, s\!=\!0) \mathcal{L}(f_i(\mathbf{x}), 1) \right. \\
& \, \left. + P(\mathbf{x}, s\!=\!0) p(y\!=\!0 \mid \mathbf{x}, s\!=\!0) \mathcal{L}(f_i(\mathbf{x}), 0) \right\}d\mathbf{x}\vphantom{\int}.
\end{aligned}
\end{equation}
\section{Relation to Other SPML Losses}\label{B}

\renewcommand{\arraystretch}{1.5}
\begin{table*}
    \centering
    \resizebox{\linewidth}{24mm}{
    \begin{tabular}{|c|c|c|c|c|c|}
        \hline
        \multirow{2}{*}{Methods} 
        & EN Loss & EM Loss & Hill Loss & Focal Margin+ SPLC & GR Loss\\
        & ~\cite{verelst2023spatial}  & ~\cite{zhou2022acknowledging} & ~\cite{zhang2021simple} & ~\cite{zhang2021simple} & (Ours)\\
        \hline
        $\hat{k}(p)$ & 
        $\begin{cases}
            0 & p \leq \tau_{1} \\
            \text{undefined} & \text{otherwise}
        \end{cases}$ & 
        $\begin{cases}
            0 & p\!\leq\!\tau_{1} \\
            1 & \text{otherwise}
        \end{cases}$ & 
        $0$ & 
        $\begin{cases}
            0 & p\!\leq\!\tau_{1} \\
            1 & \text{otherwise}
        \end{cases}$ & $\frac{1}{1 + \exp\{- (w\cdot p + b)\}}$\\
      
        \hline
        $v(p)$ & 
        $\begin{cases}
            1 & \text{otherwise} \\
            0 & s\!=\!0 \text{ and } p\!>\!\tau_{1}
        \end{cases}$ & 
        $\begin{cases}
            1 & \text{otherwise} \\
            \alpha & s\!=\!0 \text{ and } p\!>\!\tau_{1}\\
            \beta & s\!=\!0 \text{ and } p\!\leq\!\tau_{1}
        \end{cases}$ & 
        $1$ & $1$ & $\begin{cases}
    1 & s = 1, \\
    \exp\{-\frac{(p - \mu)^2}{2\sigma^2}\} & s = 0.
\end{cases}$\\
        \hline
        $\mathcal{L}_{1}$ & $-\log(p)$ & $-\log(p)$ & $-\log(p)$ & $(1-p_{m})^{\gamma} \log(p_{m})$ & $\frac{1-p^{q_1}}{q_1}$\\
        \hline
        $\mathcal{L}_{2}$ & $-\log(p)$ & $p \log(p)\!+\!(1-p) \log(1-p)$ & \text{undefined} & $(1-p_{m})^{\gamma} \log(p_{m})$ & $\frac{1-p^{q_2}}{q_2}$\\
        \hline
        $\mathcal{L}_{3}$ & $-\log(1-p)$ & $-\hat{p} \log(p)\!-\!(1-\hat{p}) \log(1-p)$ & $-(\lambda-p) p^{2}$ & $-(\lambda-p) p^{2}$ & $\frac{1-(1-p)^{q_3}}{q_3}$\\
        \hline
    \end{tabular}
}
    \caption{Representing the existing MLML/SPML losses in our unified loss function framework (Same as table \ref{t1} in the main text).}
    \label{t11}
\end{table*}

In this section, we provide more detailed explanations for unifying existing SPML/MLML methods into our loss function framework, corresponding to Section \ref{relation} and Table \ref{t1}.
To facilitate the readability, we present Table \ref{t11} in the following, which is the same as Table \ref{t1} in the main text.
\paragraph{EN Loss.}
At the beginning of each epoch $t$, for each class $i$, it identifies $N\!-\!N_i$ instances with the \emph{lowest} EMA value of label confidence $p_{i}$ and annotated them with 0 (i.e., negative pseudo labels), where $N$ is the total number of examples, and \(N_i\) is the estimated number of positive examples in the \(i\)-th class.
At the end of each epoch  $t$, the loss is recalculated with positive labels and the negative pseudo labels.
The loss can be explicitly calculated as:
\begin{equation}
\mathcal{L}_{\mathrm{EN}}\left(\mathbf{p}^t\right)\!=\!-\frac{1}{L}\!\sum_{i=1}^{L}\!\left[z_{i}\!=\!1\right] \log \left(p_{i}^t\right)\!+\!\left[\hat{z}_{i}^{t}\!=\!-1\right] \log \left(1\!-\!p_{i}^t\right),
\end{equation}
where $\hat{z}_{n i}^{t} \in\{0,1\}$, 1 and -1 indicate positive and negative labels, respectively.
In the following analysis, let \(p\) represent the label confidence \(p_{i}\).
Because existing work with pseudo-labeling strategies equivalently involve both thresholds and rankings, for convenience, we replace the ranking in a threshold manner, meaning the confidence level $\tau_{1}\in[0,1]$ of the labels on the boundary is used as the threshold.
Consequently, when \(p \leq \tau_{1}\), the label is labeled as 0, and the negative loss $\log(1\!-\!p)$, i.e., \(\mathcal{L}_3\) in the second column of Table \ref{t11}, is calculated.
In this case, \(\hat{k}(p)\!=\!0\), \(v(p)\!=\!1\) (see the second column of Table \ref{t11}).
When \(p > \tau_{1}\), the label remains a missing label, and no loss is calculated, so the loss weight \(v(p)\!=\!0\), \(\hat{k}(p)\) is undefined.
Specifically, the threshold $\tau_1$ is determined according to the pseudo-labeling strategy described above, and \(\mathcal{L}_1\), \(\mathcal{L}_2\), \(\mathcal{L}_3\) are all BCE loss.
\paragraph{EM Loss.}
The strategy of EM Loss involves computing the maximum entropy loss, i.e., $\mathcal{L}_2$ in third column of Table \ref{t11}.
For missing labels, this maximum entropy loss can be considered as a type of robust loss.
It is combined with the pseudo-labeling strategy APL \cite{zhou2022acknowledging}, which sorts the label confidence $p_c$ for each class $c$ in every epoch.
The lowest $\theta\%$ (hyperparameter) of confidence labels in each class are labeled as -1 (negative), and the negative loss $\mathcal{L}_{*}$, i.e., $\mathcal{L}_3$ in the third column of Table \ref{t11}, is calculated as:
\begin{equation}
\mathcal{L}_{*}=\hat{p}_{c} \log \left(p_{c}\right)+\left(1-\hat{p}_{c}\right) \log \left(1-p_{c}\right),
\end{equation}
where \(\hat{p}_{c} = p_{c}\), but without gradients backpropagation.
This means when calculating the derivative of $\mathcal{L}_{*}$ with respect to $p_c$, we treat \(\hat{p}_c\) as a constant.
The total loss is:
\begin{equation}\label{EM}
\begin{aligned}
\mathcal{L}_{\mathrm{EM+APL}}(\mathbf{p}, \mathbf{y})=&-\frac{1}{C} \sum_{c=1}^{C}[\mathrm{1}_{[y_{c}=1]} \log (p_{c})\\
&+\mathrm{1}_{[y_{c}=0]} \alpha H(p_{c})
+\mathrm{1}_{[y_{c}=-1]} \beta \mathcal{L}_{*}],
\end{aligned}
\end{equation}
\begin{equation}
H(p_{c})\!=\!\!-\!\left[p_{c} \log (p_{c})\!+\!(1\!-\!p_{c}) \log (1\!-\!p_{c})\right],
\end{equation}
where $\alpha$ and $\beta$ are hyperparameters.
In the following discussion, let $p$ represent the label confidence $p_c$.
Similar to the EN Loss, we use the confidence level of labels on the boundary as the threshold $\tau_{1}\in[0,1]$, thereby representing ranking uniformly through thresholds.
When $p \leq \tau_{1}$, the label is set as -1, and the negative loss is $\mathcal{L}_{*}$, i.e., $\mathcal{L}_3$ in the third column of Table \ref{t11}, is calculated.
In this case, $\hat{k}(p)\!=\!0$ and $v(p)\!=\!\beta$ (see the third column of Table \ref{t11}).
When $p\!>\!\tau_{1}$, the label remains a missing label, and the maximum entropy loss, i.e., $\mathcal{L}_2$, is computed, with the loss weight $v(p)\!=\!\alpha$ and $\hat{k}(p)\!=\!1$.
To sum up, $\tau_1$ is determined according to the pseudo-labeling strategy described above, and $\mathcal{L}_1$ and $\mathcal{L}_3$ use BCE loss, while $\mathcal{L}_2$ is the maximum entropy loss $H(p_{c})$.
\paragraph{Hill Loss.}
Hill Loss \cite{zhang2021simple} is a robust loss function proposed to address the issue of false negatives.
It is developed under the Assumed Negative (AN) assumption, which posits that all missing labels are negative.
The definition of Hill Loss is:
\begin{equation}\label{Hill}
\mathcal{L}_{Hill}^{-}=(\lambda-p) p^{2},
\end{equation}
where $\lambda$ is a hyperparameter, and in the original paper $\lambda\!=\!1.5$.
This method introduces a robust loss without employing a pseudo-labeling strategy.
Therefore, we have $\hat{k}(p)\!=\!0$, $v(p)\!=\!1$ (see the fourth column of Table \ref{t11}).
In addition, $\mathcal{L}_1$ uses the positive part of BCE loss, which is $\log(p)$.
Due to the AN assumption, $\mathcal{L}_2$ is undefined, and $\mathcal{L}_3$ is represented by $\mathcal{L}_{Hill}^{-}$.
\paragraph{Focal Margin+ SPLC.}
Self-paced Loss Correction (SPLC) is a strategy that gradually corrects potential missing labels during the training process \cite{zhang2021simple}.
It also involves determining the model's potential pseudo-labels based on the label confidence \( p \), and can be viewed as a form of online pseudo-label estimation.
The SPLC loss is defined as:
\begin{equation}
\left\{
\begin{array}{l}
\mathcal{L}_{SPLC}^{+}=loss^{+}(p), \\
\mathcal{L}_{SPLC}^{-}=\mathbb{I}(p \leq \tau)loss^{-}(p)+(1-\mathbb{I}(p \leq \tau)) loss^{+}(p).
\end{array}
\right.
\end{equation}
The parameter \(\tau\in[0,1]\) is a hyperparameter.
Although this approach is the most similar to the framework we proposed, our framework instead uses a continuous function with higher degrees of freedom, i.e., \(\hat{k}(p) \) in Eq.\eqref{eq:k}, to estimate the probability of potential positive labels online, whereas SPLC estimates through a step function.
In addition, $loss^{+}(p)$ and $loss^{-}(p)$ are decoupled positive and negative losses.
This method further addresses the false negative issue by combining the Focal margin loss \( \mathcal{L}_{\text{fml}}^{+}\!=\!(1\!-\!p_{m})^{\gamma} \log(p_{m}) \), where \( p_{m}\!=\!\sigma(x\!-\!m) \) and \( m \) is a margin parameter, along with \( \mathcal{L}_{Hill}^{-} \) given in Eq.\eqref{Hill}.
Therefore, $loss^{+}(p)$ and $loss^{-}(p)$ are defined as $\mathcal{L}_{\text{fml}}^{+}$ and $\mathcal{L}_{Hill}^{-}$, respectively.
As discussed above, in this method, \( \hat{k}(p) \) is a step function, similar to the pseudo-labeling strategy, and the weight \( v(p)\!=\!1 \).
To sum up, the losses \( \mathcal{L}_1 \) and \( \mathcal{L}_2 \) use \( \mathcal{L}_{\text{fml}}^{+} \), and \( \mathcal{L}_3 \) is \( \mathcal{L}_{Hill}^{-} \) (see the fifth column in Table \ref{t11}).
\section{Gradient Analysis}\label{C}
In this section, we demonstrate how to derive the gradient of $\mathcal{L}_{\varnothing}$, including GR Loss, EM Loss, and Hill Loss.
In addition, we provide detailed explanations for both conclusions in Section \ref{GA}. 
\subsection{GR Loss}
First, the $\mathcal{L}_{\varnothing}$ for GR Loss is known as
\begin{equation}
\mathcal{L}_{\varnothing}\!=\!\hat{k} \cdot \frac{1-p^{q_2}}{q_2}\!+\!(1\!-\!\hat{k}) \cdot \frac{1\!-\!(1\!-\!p)^{q_3}}{q_3}.
\end{equation}
Then, the gradient of $\mathcal{L}_{\varnothing}$ \emph{w.r.t.} \(z\) is given by the chain rule:
\begin{equation}
\frac{\partial \mathcal{L}_{\varnothing}}{\partial z}\!=\!\frac{\partial \mathcal{L}_{\varnothing}}{\partial p} \frac{\partial p}{\partial z}.
\end{equation}
Because $p\!=\!\sigma(z)$, we have:
\begin{equation}
\frac{\partial p}{\partial z}=\frac{e^{z}}{\left(1+e^{z}\right)^{2}}=p \cdot(1-p).
\end{equation}
Therefore, the gradient for $\mathcal{L}_{\varnothing}$ \emph{w.r.t.} $z$ becomes:
\begin{equation}
\begin{aligned}
\frac{\partial \mathcal{L}_{\varnothing}}{\partial z}\!=&\left[\hat{k}(-p^{q_{2}\!-\!1})\!+\!(1-\hat{k}) (1-p)^{q_{3}\!-\!1}\right]\!\cdot\! p\cdot(1-p)\\
=&(1-\hat{k})(1-p)^{q_3} \cdot p - \hat{k} p^{q_2} \cdot (1-p),
\end{aligned}
\end{equation}
which verifies Eq.\eqref{e18} in the main text.
\subsection{EM Loss}
According to Eq.\eqref{EM}, we obtain the EM Loss of \(\mathcal{L}_{\varnothing}\) (i.e., the loss for missing labels) as:
\begin{equation}
\mathcal{L}_{\varnothing}=p \log p+(1-p) \log (1-p).
\end{equation}
Therefore, the gradient is:
\begin{equation}
\begin{aligned}
\frac{\partial \mathcal{L}_{\varnothing}}{\partial z} & =[\log p-\log (1-p)] \cdot p\cdot(1-p) \\
& =\log \frac{p}{1-p} \cdot p\cdot(1-p).
\end{aligned}
\end{equation}
\subsection{Hill Loss}
According to Eq.\eqref{Hill}, we obtain the Hill Loss of \(\mathcal{L}_{\varnothing}\) (i.e., the loss for missing labels) as:
\begin{equation}
\mathcal{L}_{\varnothing}=(\lambda-p) p^{2}.
\end{equation}
Assuming $\lambda\!=\!1.5$, the gradient is:
\begin{equation}
\begin{aligned}
\frac{\partial \mathcal{L}_{\varnothing}}{\partial z} & =p^{2}(2 \lambda-3 p)(1-p)\\
&=3p^2(1-p)^2.
\end{aligned}
\end{equation}

\subsection{Explanation of Conclusion 1}

\paragraph{Conclusion 1 (restated).}
The imbalance between positive and negative labels is a key factor to affect SPML performance.
By properly adjusting \(q_3\), we can rebalance the supervision information. 

In SPML, each sample has only one positive label.
Suppose that there are \(C\) classes in the dataset, the \emph{positive} supervisory information constitutes only \(\frac{1}{C}\) of the total.
If the loss gradient for missing labels is equivalently weighted as that is in the positive loss gradient, e.g., using the AN loss, negative samples will dominate the gradient-based parameter update, making it difficult for the model to learn effective information from the positive samples.
In our GR loss, adjusting \(q_3\) greatly changes the gradient magnitude on the missing labels.
A larger \(q_3\) generally results in smaller gradient values, making it easier for the model to learn from the positive samples.
However, if \(q_3\) is too large, the gradient becomes too small, and the positive samples over-dominate the gradient update, similar to EPR~\cite{cole2021multi}.
Hence, it is crucial to balance the supervisory information of positive and negative samples by adjusting $q_3$ properly.

\renewcommand{\arraystretch}{1.25} 
\begin{table*}[t]
\centering
\begin{tabular}{c|c|cccc}
\toprule 
\multicolumn{2}{c|}{Statistics} & VOC & COCO & NUS & CUB \\ 
\midrule 
\multicolumn{2}{c|}{\# Classes} & 20 & 80 & 81 & 312\\
\midrule
\multirow{3}{*}{\# Images}
& Training & 4,574 & 65,665 & 120,000 & 4,795 \\
& Validation & 1,143 & 16,416 & 30,000 & 1,199 \\
& Test & 5,823 & 40,137 & 60,260 & 5,794 \\
\midrule 
\# Labels Per  & Positive & 1.46 & 2.94 & 1.89 & 31.4 \\
Training Image & Negative & 18.54 & 77.06 & 79.11 & 280.6 \\
\bottomrule
\end{tabular}
\caption{Statistics of the datasets, including the number of classes, the number of images on the split datasets, and the averaged number of ground-truth positive and negative labels per image on the training sets.}
\label{tab:Datasets}
\end{table*}
\begin{figure*}[t!]
    \centering
    \begin{subfigure}[t]{0.44\textwidth}
           \centering
           \includegraphics[width=\textwidth]{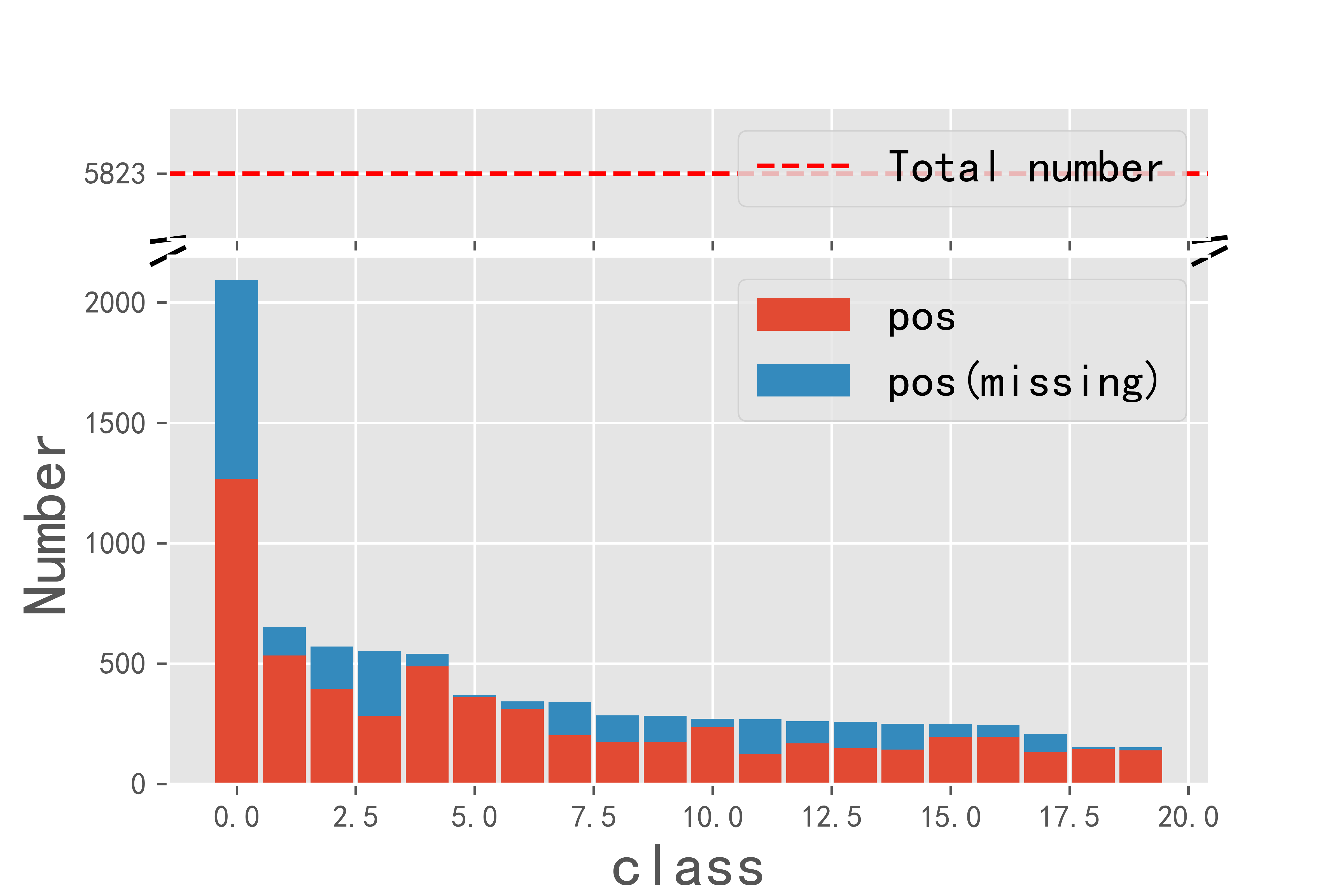}
            \caption{VOC}
            \label{fig:voc.png}
    \end{subfigure}
    \begin{subfigure}[t]{0.44\textwidth}
            \centering
            \includegraphics[width=\textwidth]{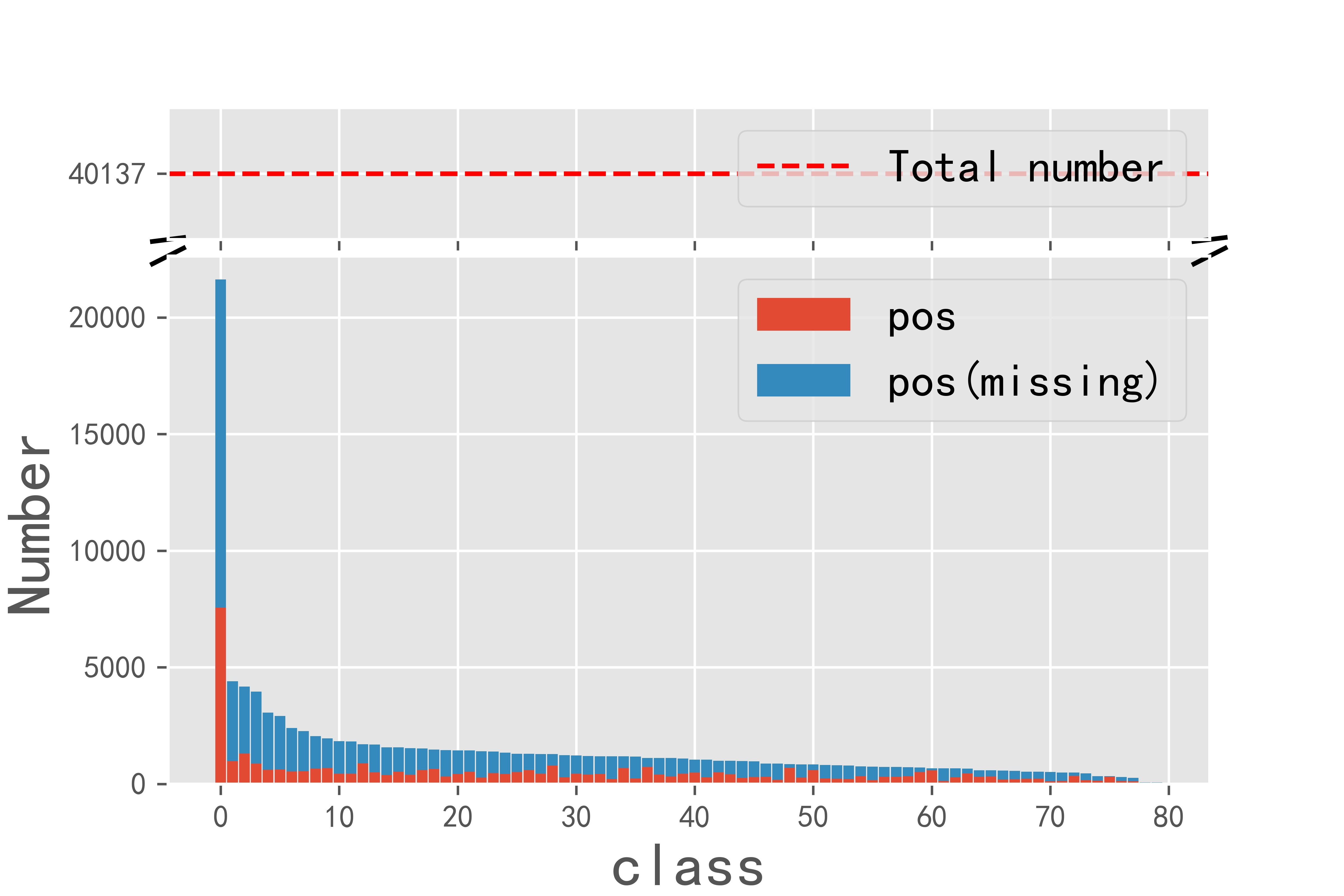}
            \caption{COCO}
            \label{fig:coco.png}
    \end{subfigure}
    \\
    \begin{subfigure}[t]{0.44\textwidth}
            \centering
            \includegraphics[width=\textwidth]{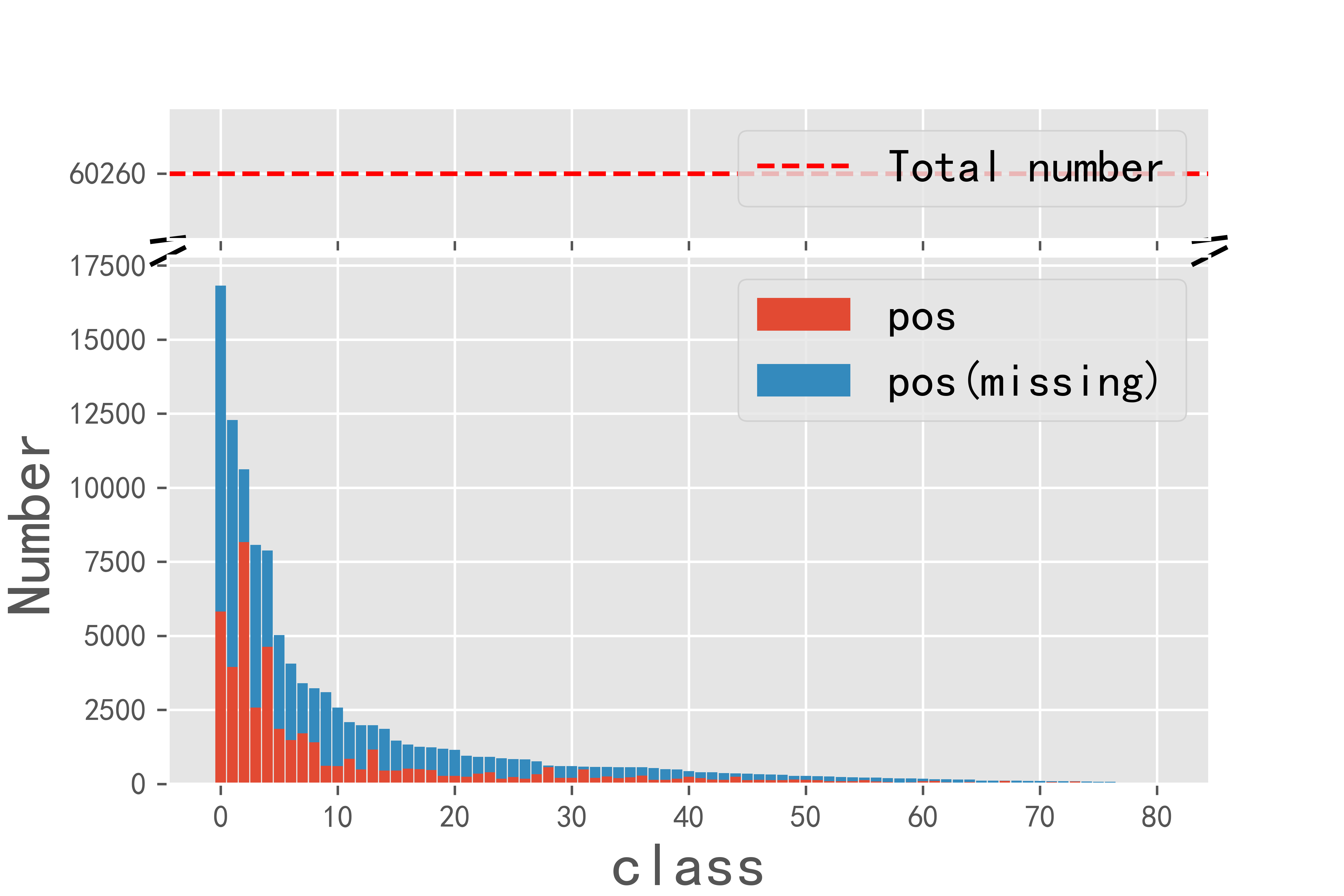}
            \caption{NUS}
            \label{fig:nus.png}
    \end{subfigure}
    \begin{subfigure}[t]{0.44\textwidth}
            \centering
            \includegraphics[width=\textwidth]{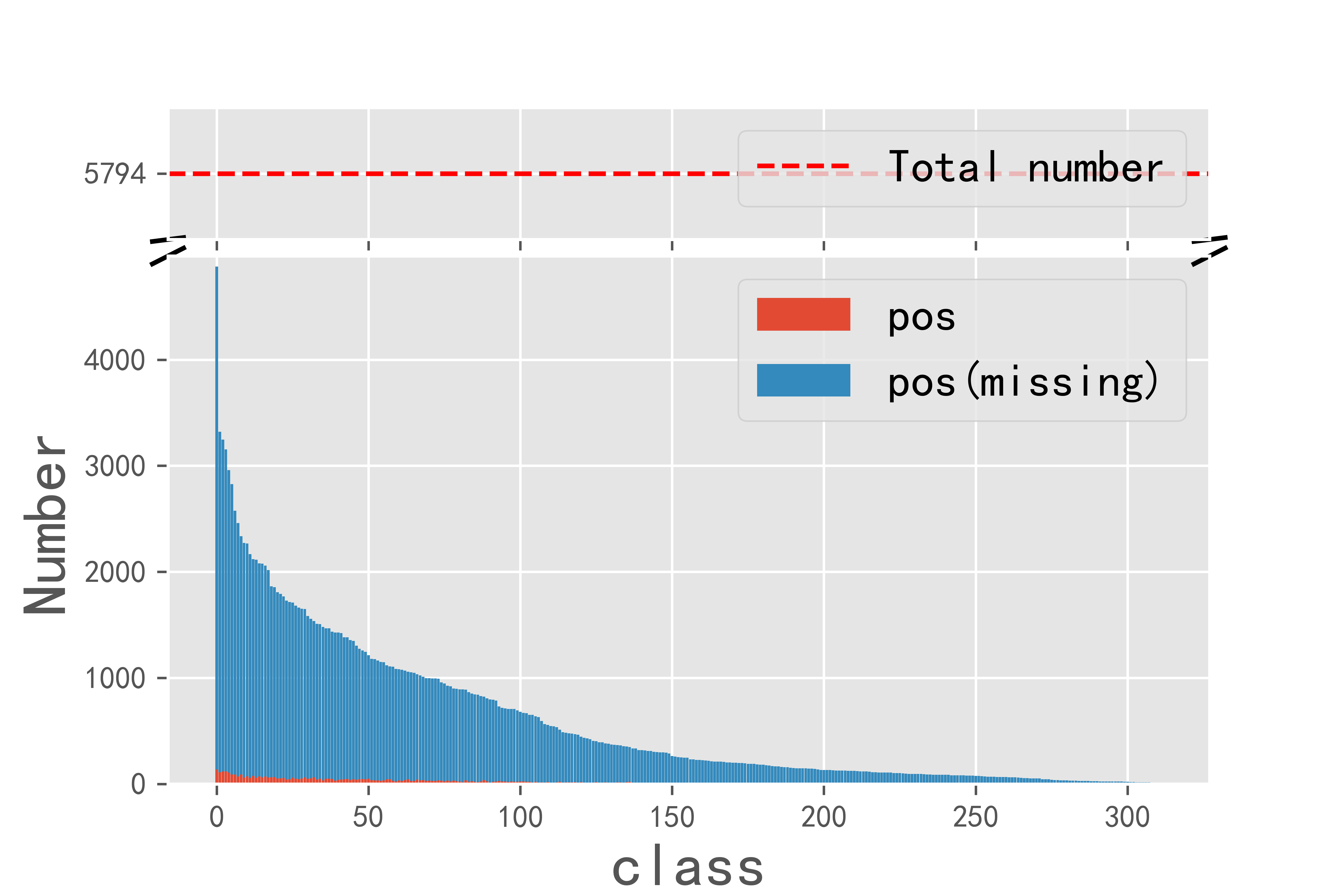}
            \caption{CUB}
            \label{fig:cub.png}
    \end{subfigure}
    \caption{The numbers of different samples \emph{w.r.t.} the class. The number of positive labels (i.e., pos), the number of positive labels among the missing labels, i.e., pos (missing), and total number of samples represented by a dashed line for each class in four datasets.}
    \label{fig:Datasets}
\end{figure*}

\renewcommand{\arraystretch}{1.25} 
\begin{table*}[t]
\centering
\begin{tabular}{c|cccc}
\toprule 
\textbf{hyperparameters} & \textbf{VOC} & \textbf{COCO} & \textbf{NUS} & \textbf{CUB} \\ 
\midrule 
Batch Size & 8 & 16 & 16 & 8 \\
Learning Rate & 1e-5 & 1e-5 & 1e-5 & 5e-5 \\
Epoch (T) & 8 & 8 & 10 & 10 \\
$q_2$ & 0.01 & 0.01 & 0.01 & 0.01 \\
$q_3$ & 1 & 1 & 1 & 1.5 \\
$\beta^{(T)}=(w^{(T)},b^{(T)})$ & (2, -2) & (10, -8) & (10, -8) & (10, -8) \\
$\alpha^{(T)}=(\mu^{(T)},\sigma^{(T)})$ & (0.8, 0.5) & (0.8, 0.5) & (0.8, 0.5) & (0.8, 0.5) \\
\bottomrule
\end{tabular}
\caption{Hyperparameter settings of our GR loss on all the four benchmark datasets in the experiments.}
\label{tab:hyperparameters}
\end{table*}

\begin{figure*}[t!]
    \centering
    \begin{subfigure}[t]{0.4\textwidth}
           \centering
           \includegraphics[width=\textwidth]{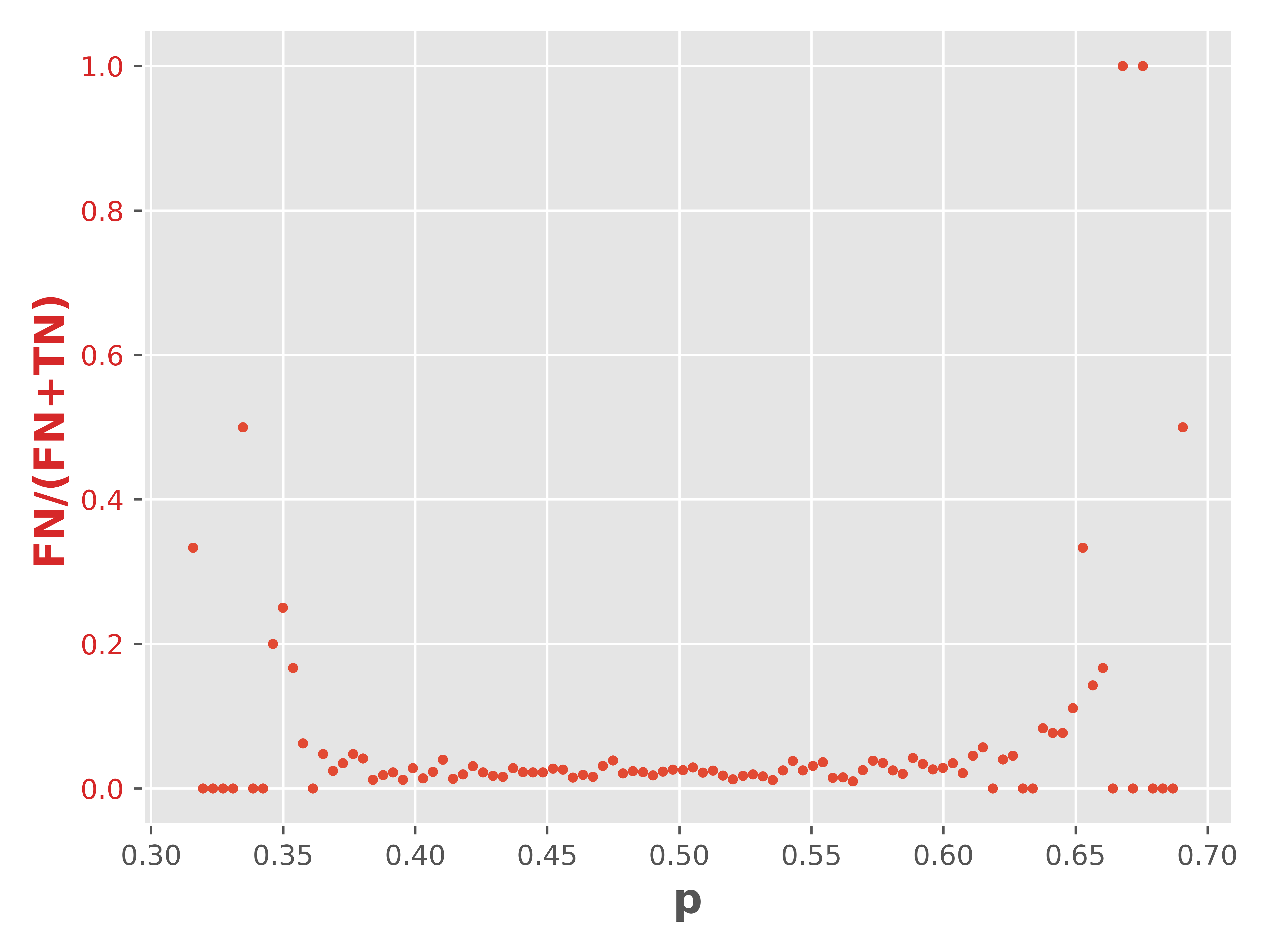}
            \caption{The beginning of training on VOC}
            \label{fig:voc0}
    \end{subfigure}
    \begin{subfigure}[t]{0.4\textwidth}
            \centering
            \includegraphics[width=\textwidth]{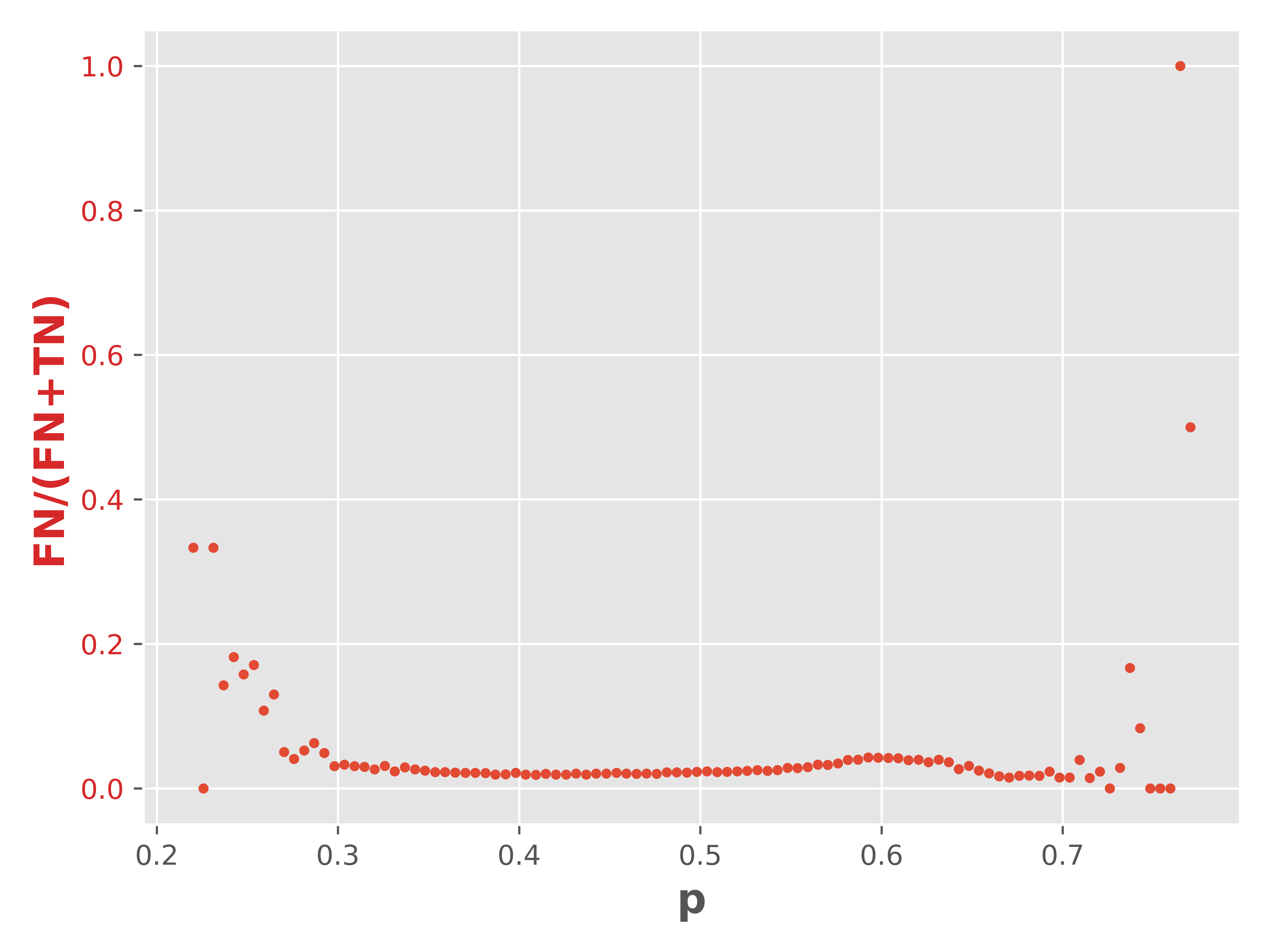}
            \caption{The beginning of training on COCO}
            \label{fig:coco0}
    \end{subfigure}
    \\
    \begin{subfigure}[t]{0.44\textwidth}
            \centering
            \includegraphics[width=\textwidth]{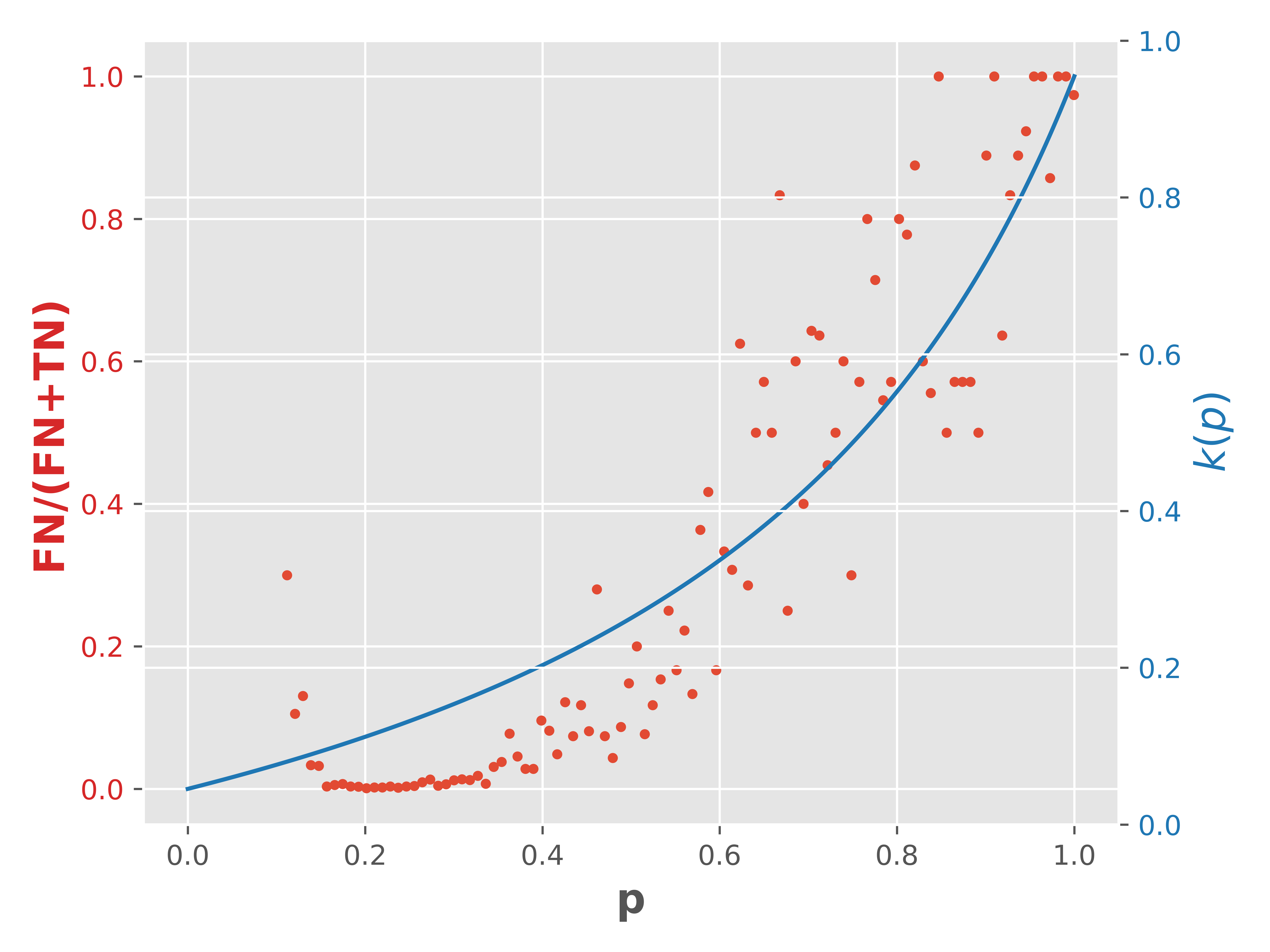}
            \caption{The end of training on VOC}
            \label{fig:voc1}
    \end{subfigure}
    \begin{subfigure}[t]{0.44\textwidth}
            \centering
            \includegraphics[width=\textwidth]{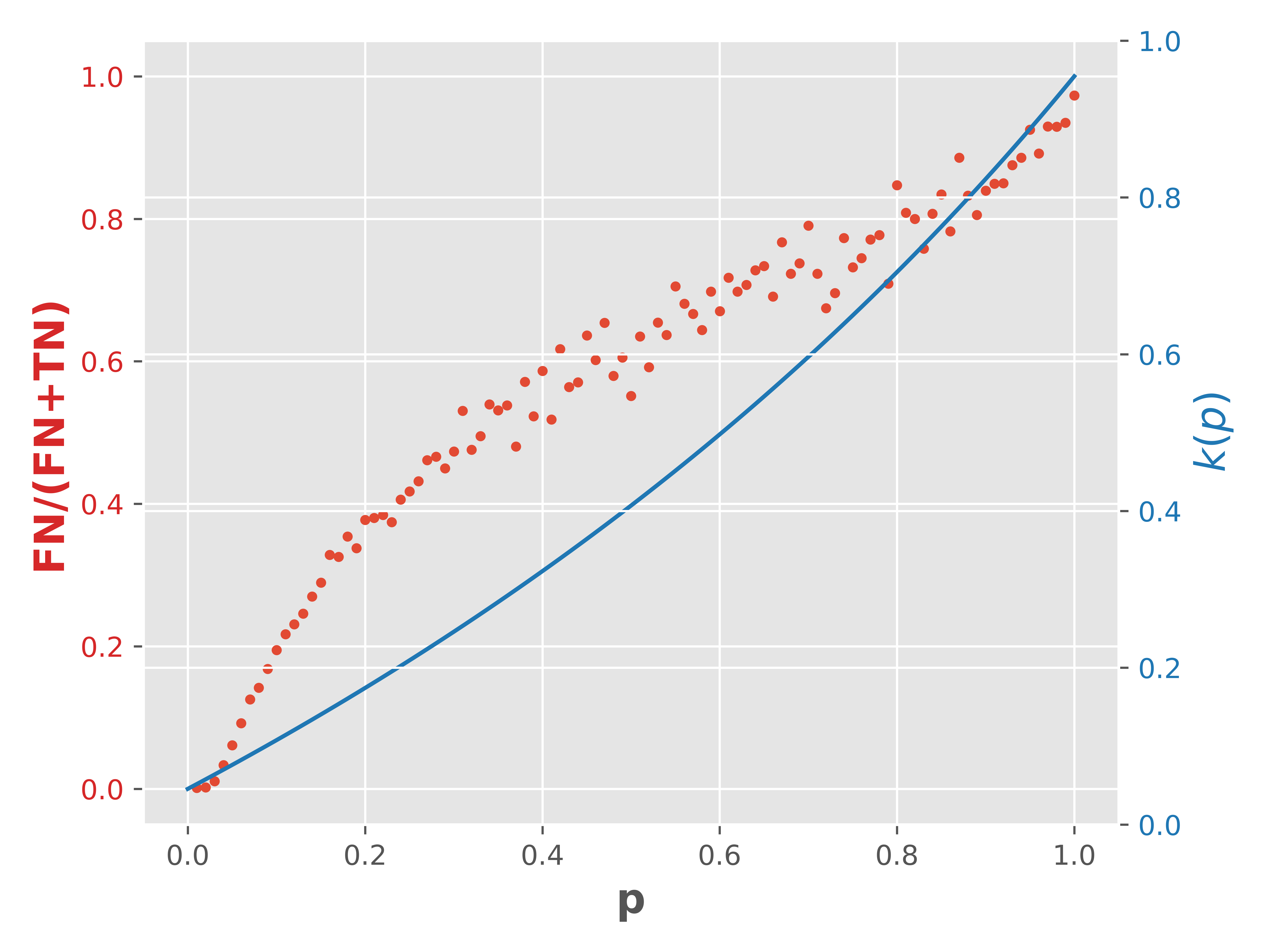}
            \caption{The end of training on COCO}
            \label{fig:coco1}
    \end{subfigure}
    \caption{The average values of FN/(FN+TN) across 100 intervals on the VOC and COCO. (a) represents the beginning of training on VOC, (b) is the beginning of training on COCO, (c) illustrates the end of training along with $\hat{k}(p)$ curve (the blue one) in Eq.\eqref{kx}, where $a\!=\!1/1.46$, and (d) displays the end of training along with $\hat{k}(p)$ curve (the blue one) in Eq.\eqref{kx}, where $a\!=\!1/2.94$.
    }
    \label{fig:ev}
\end{figure*}

\subsection{Explanation of Conclusion 2}

\paragraph{Conclusion 2 (restated).}

The issue of false negatives is also a key factor to impact the gradient (see Eq.\eqref{e18}), which can be effectively addressed by adjusting \(\hat{k}(p;\beta)\).

First, our experiments show that AN loss does not adequately address the issue of false negatives, as shown in Figure (\ref{fig:AN}), where there are still many false negatives near \(p = 0\). 
In contrast, for the GR loss, at the beginning of training, as shown in the curve of \(\beta^{(0)}\) (the blue one) in Figure (\ref{fig:gradient}), when \(p\) is small, the gradient is negative.
Hence, \(p\) will be increased along the training so as to decrease the loss.
In the early stage of training, because the model is not well-trained, this can increase the confidence \(p\) of all the samples, thus avoid to put low confidence to the false negatives.

On the other hand, in the late stage of training, as shown in the curve of \(\beta^{(T)}\) (the green one) in Figure (\ref{fig:gradient}), for a missing label, when \(p\) is close to 1, it is more likely the corresponding sample to be a false negative.
Note that the gradient is also negative, so that its confidence \(p\) will be further increased.

\section{Robust Loss}\label{D}
In Section \ref{Lq}, we introduce a robust loss function based on \cite{zhang2018generalized} as:
\begin{equation}
\mathcal{L}_q = \frac{1-p^{q}}{q}.
\end{equation}
The gradient of loss \(\mathcal{L}_{q}\) is calculated as:
\begin{equation}\label{e35}
\frac{\partial \mathcal{L}_{q}\left(f\left(\mathbf{x} ; \boldsymbol{\theta}\right)\right)}{\partial \boldsymbol{\theta}}=-f(\mathbf{x} ; \boldsymbol{\theta})^{q-1} \nabla_{\boldsymbol{\theta}} f(\mathbf{x} ; \boldsymbol{\theta}).
\end{equation}
\begin{equation}\label{e36}
\frac{\partial \left(-\log\left(f\left(\mathbf{x} ; \boldsymbol{\theta}\right)\right)\right)}{\partial \boldsymbol{\theta}}=-f(\mathbf{x} ; \boldsymbol{\theta})^{-1} \nabla_{\boldsymbol{\theta}} f(\mathbf{x} ; \boldsymbol{\theta}),
\end{equation}
where \(f(\mathbf{x}; \boldsymbol{\theta})\) and \(q\) both lie in $[0,1]$.
In Eq.\eqref{e35}, \(\mathcal{L}_{q}\) loss adds an extra weight of \(f(\mathbf{x} ; \boldsymbol{\theta})^{q}\) to each sample, compared to BCE in Eq.\eqref{e36}.
This means we put less focus on samples where the softmax outputs and labels are not closely aligned, enhancing robustness against noisy data.

Compared to MAE, this weighting of \(f(\mathbf{x} ; \boldsymbol{\theta})^{q-1}\) on each sample helps to learn the classifier by paying more attention to difficult samples where the labels and softmax outputs do not match.
A larger \(q\) results in a loss function that is more robust to noise.
However, a very high \(q\) can make the optimization process harder to converge.
Therefore, as shown in Figure (\ref{fig:loss}) and Figure (\ref{fig:q}), properly setting \(q\) between 0 and 1 can strike for the promising balance between noise resistance and effective learning.

\section{Experiments}\label{E}
\subsection{Datasets Details}\label{E1}

The following four common multi-label datasets are used in our experiments.

1) PASCAL VOC 2012 (VOC)~\cite{everingham2012pascal}: Containing 5,717 training images and 20 categories.
We report the test results on its official validation set with 5,823 images.

2) MS-COCO 2014 (COCO)~\cite{lin2014microsoft}: Containing 82,081 training images and 80 categories.
We also report the test results on its official validation set of 40,137 images.

3) NUS-WIDE (NUS)~\cite{chua2009nus}: Consisted of 81 classes, containing 150,000 training images and 60,260 testing images collected from Flickr.
Instead of re-crawling NUS images as in the literature~\cite{cole2021multi}, we used the official version of NUS in our experiments, which reduces human intervention and thus much fairer.

4) CUB-200-2011 (CUB)~\cite{wah2011caltech}: Divided into 5,994 training images and 5,794 test images, consisting of 312 categories (i.e., binary attributes of birds).

For reference, we show the statistical details of these datasets in Table \ref{tab:Datasets}.
We display the count of positive labels (i.e., $\#{\{s=1\}}$) and the number of positive labels among missing labels (i.e., $\#{\{y=1,s=0\}}$)  in each class, as defined in Section \ref{define}, using bar charts to present this data in Figure \ref{fig:Datasets}. This illustrates the imbalance between positive and negative samples, as well as the issue of false negatives.

\subsection{Hyperparameter Settings}\label{E2}
In our approach, we need to determine four hyperparameters in $(\beta^{(T)}$, $\alpha^{(T)})$, where $\beta^{(T)}\!=\!(w^{(T)},b^{(T)})$, $\alpha^{(T)}\!=\!(\mu^{(T)},\sigma^{(T)})$.
Moreover, we fix the hyperparameters $q_2$ and $q_3$ for each dataset, typically 0.01 and 1, respectively. 
For each dataset, we separately determine these hyperparameters  by selecting the ones with the best mAP on the validation set for the final evaluation.
For convenience, the final hyperparameters of our method are shown in Table \ref{tab:hyperparameters}, together with the selected batch sizes and learning rates.

\subsection{Experiment to Verify Assumptions}\label{E3}
For VOC and COCO validation sets, we evenly divide $[0,1]$ into 100 intervals.
In each interval, we simplify the output label confidence \(p\) before the first epoch and after the last epoch, through a bucketing operation to calculate the ratio of false negatives to missing labels, i.e., the sum of false negatives and true negatives, denoted as FN/(FN+TN).
The result is shown in Figure \ref{fig:ev}.
This ratio reflects the probability $P(s\!=\!1 \mid y\!=\!1, \mathbf{x})$ in Eq.\eqref{kx}.
By plotting the ratio of each interval on a scatter plot, we can discern the essential characteristics that \(\hat{k}(p)\) should follow.

The results indicate that at the beginning of training, the scatter plot resembles a constant function, consistent with Assumption \ref{as1}.
At the end of training, the scatter plot appears as a monotonically increasing function, closely aligning with the function of \(\hat{k}(p)\) described in Eq.\eqref{kx}, consistent with Assumption \ref{as2}.

\end{document}